\RequirePackage{amsmath}
\RequirePackage{fix-cm}
\RequirePackage{cite}

\documentclass{article}
\usepackage[a4paper,margin=2cm]{geometry}

\usepackage{lineno}
\modulolinenumbers[1]
\usepackage{amssymb,amsmath,mathtools}
\usepackage{lmodern}
\usepackage{graphicx}
\usepackage{float}
\usepackage{bm,upgreek}% bold math
\usepackage{grffile}
\usepackage{etoolbox}
\usepackage{authblk}
\usepackage{xcolor}
\usepackage{xspace}
\usepackage{siunitx}
\usepackage{comment}
\usepackage{orcidlink}
\usepackage{booktabs}
\usepackage{longtable}
\usepackage{multirow}
\usepackage{pdflscape}
\usepackage{array}
\usepackage{makecell}
\usepackage{caption}
\usepackage{algorithmic}
\usepackage{textcomp}
\usepackage{pdfpages}
\usepackage{placeins}
\usepackage{setspace}
\PassOptionsToPackage{hyphens}{url}
\usepackage{hyperref}

\makeatletter
\def\@lbibitem[#1]#2{%
  \item\if@filesw
    {\def\protect\@noexpand{\protect\noexpand}%
     \immediate\write\@auxout{\string\bibcite{#2}{\the\value{\@listctr}}}}%
  \fi
  \ignorespaces}
\makeatother

\let\epsilon\varepsilon

\newcommand{\PMten}{PM$_{10}$\xspace}
\newcommand{\PMtwofive}{PM$_{2.5}$\xspace}

\newcommand{\SOtwo}{SO$_2$\xspace}
\newcommand{\Othree}{O$_3$\xspace}
\newcommand{\CO}{CO\xspace}
\newcommand{\NOtwo}{NO$_2$\xspace}

\newcommand{\umg}{\ensuremath{\upmu \text{g}/\text{m}^3}}
\usepackage{xcolor}

\begin{document}

\title{Leveraging Remote Traffic Data for Local Air Pollutant Estimation:
  A Scenario-Based Machine Learning Study Across London Monitoring Sites}

\author[1,2]{Valeria Legaria-Santiago\,\orcidlink{0009-0008-4843-4971}}
\author[2]{Amadeo Arguelles\,\orcidlink{0000-0001-8627-4739}}
\author[2]{Magdalena Saldana-Perez\,\orcidlink{0000-0002-2475-1621}}
\author[1]{Jocelyn Richardson\,\orcidlink{0009-0001-4147-7011}}
\author[1]{Marcella Bona\,\orcidlink{0000-0002-9660-580X}}

\affil[1]{Department of Physics and Astronomy, Queen Mary University of London, United Kingdom}
\affil[2]{Centro de Investigacion en Computacion, Instituto Politecnico Nacional, Ciudad de Mexico, Mexico}

\date{Submitted to MDPI Atmosphere journal, accepted: 17 August 2026}

\clearpage
\maketitle

\begin{abstract}
Vehicular traffic is a major source of air pollution; however, the contribution of remotely acquired traffic information to local machine-learning (ML) air-pollution models remains insufficiently characterised. This study evaluates four interpretable tree-based ML models (Random Forest, Extra Trees, LightGBM, and XGBoost) under six predictor scenarios combining progressively larger predictor sets, ranging from remotely acquired traffic, meteorological, and temporal variables alone to the inclusion of measurements from one and four neighbouring monitoring stations, to estimate NO$_2$, PM$_{10}$, PM$_{2.5}$, and O$_3$ concentrations across several sites in London. ML model performance was compared with a ridge linear regression model as a baseline, with spatial interpolation methods and with a cross-site validation experiment. When modelling without data from neighbouring stations, the RMSE for \NOtwo ranged from 9.73 to 11.66 $\upmu$g/m$^3$ without  traffic information, compared with 8.72 to 11.52 $\upmu$g/m$^3$ when traffic information was included.
Additionally, for \NOtwo, SHAP analyses indicate that traffic-related variables can contribute at levels comparable to pollutant measurements from neighbouring monitoring stations in traffic-dominated~environments.
\end{abstract}

%\keyword{air pollution prediction; machine learning; road traffic data; traffic-related air pollution; urban monitoring networks; virtual sensing} 

\section{Introduction}
\label{sec:Introduction}
Air pollution is a major health issue, often affecting more deprived areas of cities~\cite{londonAQdeprivation}. According to the World Health Organization (WHO) and the European Environment Agency (EEA), 9 of 10 people, especially those living in urban areas, are exposed to harmful levels of pollutants that do not meet the WHO air quality guidelines~\cite{oms_guidelines_2021, WHO_2022, eea2023, WHO_2025}. One strategy to tackle air pollution involves monitoring air pollutants to assess compliance with air quality standards~\cite{CityLondon2025_AirQualityStrategy}, with~particular emphasis on the six criteria pollutants defined by the United States Environmental Protection Agency~\cite{CriteriaPollutants_NAAQS}. The~criteria pollutants are ozone (\Othree), particulate matter with diameters below 10 and 2.5 micrometres (\PMten and \PMtwofive, respectively), carbon monoxide (\CO), sulphur dioxide (\SOtwo), nitrogen dioxide (\NOtwo), and~lead (Pb)~\cite{WHO_criteria_pollutants, EPA_Criteria_Air_Pollutants}.

Vehicular traffic is one of the main sources of CO, \PMten and \PMtwofive, and~emits volatile organic compounds (VOCs) and \NOtwo, which contribute to the photochemical formation of  \Othree. For this reason, several studies have incorporated traffic conditions into machine learning (ML) models for air pollution estimation. Traffic-related predictors commonly include vehicle counts, traffic speed, traffic flow, traffic volume, traffic density, and~road congestion~\cite{Du2025, LESNIK2019, Zalakeviciute2020, DEVEER2025, azeez_modeling_2019, kaminska_random_2019, KAMINSKA_2025, Sulaimon2022, ramentol_short-term_2023, CCTV_NO2_2025, Samad2023, bona2026tacklingairqualitysapiens}. In~addition to traffic-related variables, some studies have also incorporated meteorological variables that are strongly associated with pollutant formation, dispersion, and~accumulation~\cite{Liu2020AirPollutionMeteorology, taheri2024met_var_in_Ozone_studies}, including temperature, wind speed, precipitation, relative humidity, and~atmospheric~pressure.

To obtain these data, some studies have developed custom data acquisition systems, reporting promising results for the estimation of atmospheric pollutant concentrations~\mbox{\cite{LESNIK2019, Zalakeviciute2020, DEVEER2025}} and traffic-related emissions~\cite{azeez_modeling_2019}. Nevertheless, these studies typically considered a single site, limiting the generalisability and scalability of their approaches over large urban areas due to the high cost of deploying and maintaining dense monitoring networks. To~overcome these limitations and improve spatial coverage and model generalisation, more recent studies have leveraged publicly available data from fixed air quality and meteorological monitoring stations~\cite{virtual_monitoring_stations}, combined with large-scale traffic datasets obtained either from city-wide camera networks~\cite{kaminska_random_2019, KAMINSKA_2025, CCTV_NO2_2025} or from organisations providing traffic information services~\cite{Sulaimon2022, Samad2023}.

However, there is no clear consensus regarding the extent to which changes in vehicular traffic affect air quality, as~the relationship varies across locations and pollutants~\cite{wang_network_2022, ADAM2021covid, venter2020covid, gla2020airquality, Rossi2020TrafficAirPollution, lowemissionSouthKorea2025}.
For example, ref.~\cite{Rossi2020TrafficAirPollution} found that reductions in traffic flow during the COVID-19 lockdown were associated with significant decreases in NO, \NOtwo, and~NO$_x$ concentrations, whereas no significant association was found for \PMten. Similarly, the~studies by refs.~\cite{venter2020covid, gla2020airquality} reported that although decreases in \NOtwo concentrations were widespread following reductions in traffic activity, the~responses of \Othree and particulate matter were more variable and, in~some cases, even showed increases. These findings highlight the need to better understand the conditions where traffic information could improve air pollutant predictions.

To investigate the relationship between traffic and air pollution, ref.~\cite{DEVEER2025} developed five datasets containing different combinations of meteorological variables, vehicle counts (total and by vehicle type), and~vehicle emissions. They trained multilinear regression, Random Forest (RF), and~eXtreme Gradient Boosting (XGBoost) models to estimate \NOtwo, \PMten, and~\PMtwofive concentrations at a roadside location in Canada.
Although RF and XGBoost achieved the best overall performance, the~study relied on data from a single monitoring location and the traffic dataset covered only approximately eight days.
On the other hand, ref.~\cite{Sulaimon2022} compared the performance of 40 ML models trained to predict \Othree, \NOtwo, \PMten, and~\PMtwofive concentrations at 338 locations across the United Kingdom (UK), identifying XGBoost, Random Forest, Light Gradient Boosting Machine (LGBM), Bagging, Extra Trees, and~Histogram Gradient Boosting as the top six performing regression algorithms. The~models were trained locally using two datasets covering six months of observations: one containing only meteorological variables and another combining meteorological and traffic data. The~results were reported as aggregate performance metrics across all locations, showing a consistent improvement when traffic data were included. However, the~primary objective of the study was to compare model performance rather than to investigate how the contribution of traffic-related information varies across locations, potentially masking local variability.
In contrast, ref.~\cite{Samad2023} averaged the results of five models trained using meteorological and temporal variables alone against models that additionally incorporated traffic data and pollutant measurements from neighbouring stations.
Their results suggested that neighbouring-station pollutant measurements are essential for achieving accurate predictions, whereas the inclusion of traffic variables improved performance in only one of the two analysed locations, and~only for \NOtwo, but~not for \PMten or \PMtwofive. Although~the study used nearly four years of data, it considered only two locations and did not explicitly quantify the independent contribution of traffic-related information when multiple neighbouring stations were~available.

Taken together, these studies demonstrate the potential value of traffic-related information for air pollution prediction. However, the~extent to which traffic variables contribute to predictive performance across pollutants and locations remains insufficiently understood.
Motivated by these limitations, and~seeking to balance spatial coverage, data requirements, and~predictive performance, this work leverages existing air pollution monitoring data from public sources together with traffic and meteorological information from commercial providers, accessed through standard web/API services. To~this end, this study evaluates how accurately concentrations of \Othree, \NOtwo, \PMten, and~\PMtwofive can be estimated using only traffic, temporal, and~meteorological data, and~compares the results against models that additionally incorporate observations from neighbouring stations. The different predictor set combinations (referred to as scenarios) are designed to assess whether traffic information can complement existing monitoring infrastructure or support air-quality assessment at locations where monitoring stations are unavailable.
The~comparisons are made both with and without traffic-related variables. Furthermore, we investigate whether the contribution of traffic-related information is consistent across locations or depends on site-specific~characteristics.

The air quality monitoring network in London includes different station types designed to characterise pollution associated with different emission sources. This study focuses on traffic-type monitoring stations (also referred to as traffic stations) as target locations, where the contribution of traffic-related information to pollutant estimation is expected to be more evident due to the influence of nearby road traffic. Background-type stations, which are located away from dominant local emission sources and where the sampled air represents a mix of all the sources within an area of several kilometres around, are used as complementary sources of pollutant measurements.

By selecting traffic stations as target locations, the~models provide contemporaneous local-scale pollutant estimates representative of traffic-influenced conditions along the road segment where each station is located. According to the monitoring-site specification~\cite{environment_type}, air sampled at traffic stations is representative of a street segment of at least 100 m in length.

This study focuses exclusively on tree-based machine learning models (RF, Extra Trees, LightGBM, and~XGBoost), as~these algorithms have consistently achieved strong predictive performance in similar studies.
Moreover, their interpretability enables the contribution of individual features to model predictions to be quantified. In this study, whether the inclusion of traffic-related variables improves predictive performance is first assessed through ablation, after which feature importance is used to quantify their contribution to the predictions of those models.
Model performance is also compared against interpolation methods, namely Kriging and Inverse Distance Weighting (IDW), which rely solely on measurements from nearby monitoring stations and are used as spatial baselines, as~well as against Ridge regression as a linear regression baseline.

%%%%%%%%%%%%%%%%%%%% here
\section{Materials and~Methods}
\label{sec:Methods}
\unskip
\subsection{Data Collection and~Pre-Processing}

The study period spanned from 23 June 2025 to 12 December 2025, comprising almost six months of concurrent hourly traffic, meteorological, and~air quality~observations.

The target pollutants are \NOtwo, \PMtwofive, \PMten, and~\Othree, estimated at the locations of the London Marylebone Road, Camden Kerbside, and~Wandsworth–Putney High Street monitoring stations. Model predictions are subsequently compared against the pollutant concentrations measured at the corresponding stations. All atmospheric pollutant data are obtained from the publicly available data archive of the Department for Environment, Food \& Rural Affairs via the UK-AIR website~\cite{defra_ukair_data} and consist of hourly mean concentrations reported in units of \umg\ using GMT hour-ending~timestamps.

\begin{figure}[H]
  \centering
    \includegraphics[width=0.79\linewidth]{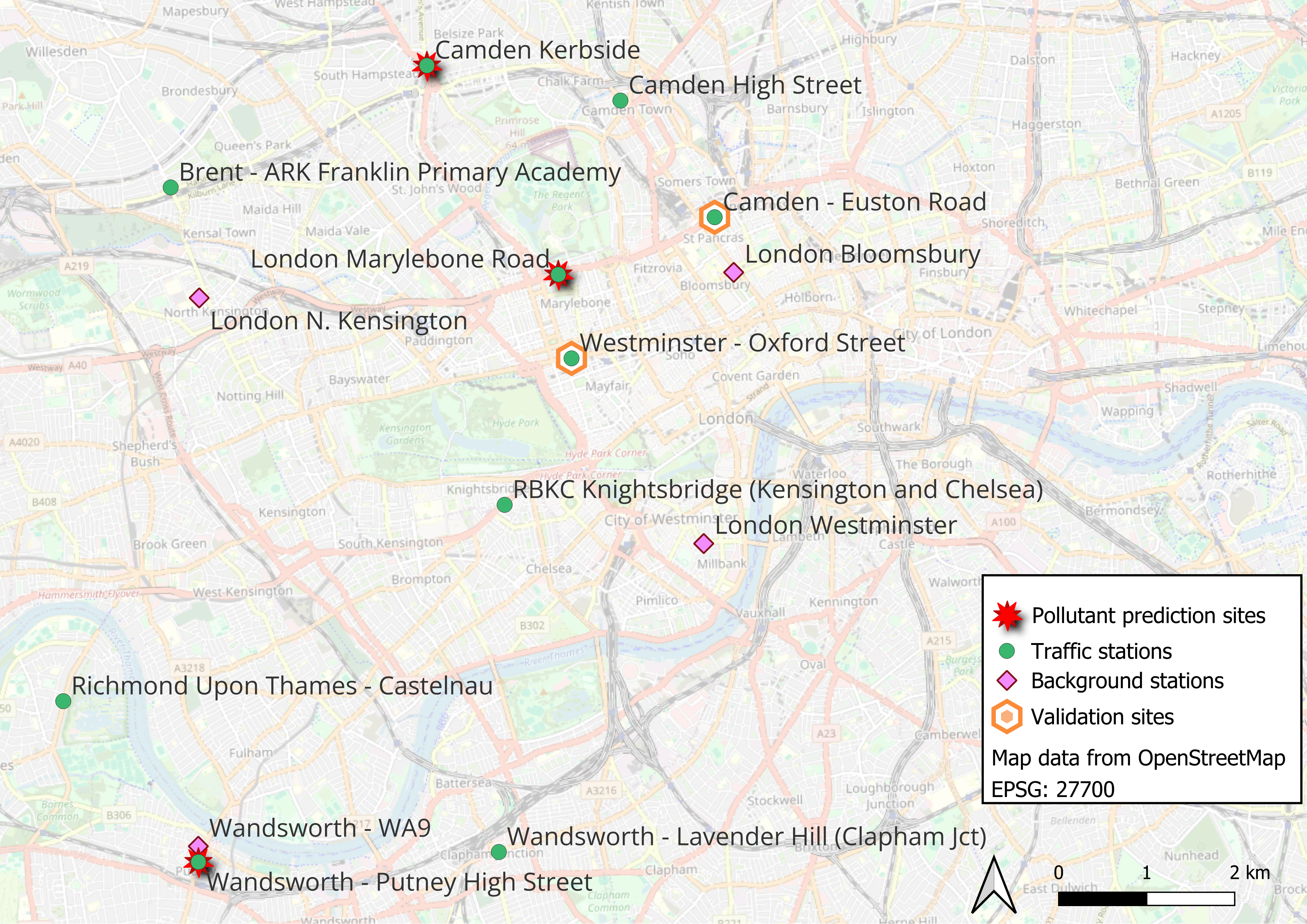}
    \caption{Locations of the pollutant prediction sites and the validation sites (all corresponding to traffic-type monitoring stations), together with the auxiliary stations (including both background and traffic-type stations).}
    \label{fig:Map_stations}
\end{figure}

Figure~\ref{fig:Map_stations} shows the target sites for air pollutant prediction as red star markers. These sites correspond to traffic-type monitoring stations, and~their data are used to train the machine learning models using supervised learning. Traffic-type stations (represented by green circle markers on the map) are located in areas where pollution levels are predominantly influenced by nearby traffic emissions, and~the sampled air is representative of at least a 100~m road segment~\cite{environment_type}.

Air pollutant measurements from nearby traffic-type and background-type stations (hereafter referred to as neighbouring stations) are used as additional predictors in the machine learning models (see Section~\ref{sec:feature_combination} and Table~\ref{tab:distance_matrix_between_stations}). Background-type stations (represented by pink diamond markers on the map) are located in areas where pollution is not dominated by a specific source or street, but~instead reflects contributions from multiple emission sources within a radius of several~kilometres.

\begin{table}[H]
  \scriptsize
    \caption{Distances between the target locations used in the main and validation analyses and their corresponding neighbouring background and traffic-type~stations.}
    \label{tab:distance_matrix_between_stations}
    \newcolumntype{L}[1]{>{\raggedright\arraybackslash}p{#1}}
    \hspace{-1.2cm}
    \begin{tabular}{L{2.5cm}L{2.5cm}L{2.5cm}L{4cm}L{2.5cm}L{2.8cm}}
        \toprule
        \textbf{Distance Between Stations }& \textbf{Camden Kerbside} & \textbf{London Marylebone Road} & \textbf{Wandsworth--Putney High Street} & \textbf{Camden--Eusto Road} & \textbf{Westminster--Oxford Street}\\
        \midrule
        1st nearest background station & London N. Kensington (3.69 km) & London Bloomsbury (1.99 km) & Wandsworth--Putney (0.18 km) & London Bloomsbury (0.66 km) & London Bloomsbury (2.08~km)\\
        \midrule
        2nd nearest background station & London Bloomsbury (4.20 km) & London Westminster (3.47 km) & London N. Kensington (6.41 km) & London Westminster (3.71 km) & London Westminster (2.58~km) \\
        \midrule
        1st nearest traffic station & Camden High Street (2.23 km) & Camden High Street (2.10 km) & Richmond Upon Thames--Castelnau (2.38 km)  & Camden High Street (1.70 km) & Camden High Street (2.98~km)\\
        \midrule
        2nd nearest traffic station & Brent--ARK Franklin Primary Academy (3.22~km) & RBKC Knightsbridge (2.69~km) & Wandsworth--Lavender Hill (3.41~km) & RBKC Knightsbridge (4.04~km) & RBKC Knightsbridge (1.83~km)\\
        \bottomrule
    \end{tabular}
\end{table}

Camden–Euston Road and Westminster–Oxford Street, both classified as traffic-type stations, serve as validation sites (represented by orange hexagonal markers in Figure~\ref{fig:Map_stations}). Their data are excluded from model training and all primary analyses and are used exclusively during the final external validation stage of the~study.

Since traffic-type stations are designed to characterise roadside emissions from nearby roads, traffic-related data are collected from the road segment closest to each target location. These road segments correspond to Marylebone Road, Putney High Street, and~Finchley Road, the~latter being the nearest road to the Camden Kerbside station.
Marylebone Road is a major arterial road in Central London, configured as a multi-lane (six-lane) two-way carriageway. It carries high traffic volumes while remaining adjacent to large green areas.
Putney High Street is a narrow two-lane urban street with one lane in each direction. It functions as a local commercial corridor with moderate traffic flow and frequent pedestrian activity.
Finchley Road is a primary route connecting Central and North London, characterised by multiple lanes in each direction and substantial traffic density. It serves both local and through traffic, with~a mix of residential, commercial, and~transit-related activity along its~corridor.

Traffic data are collected in real time every 5 min from the TomTom platform using the Flow Segment Data service~\cite{TomTom_Flow_Segment_Data} through the free plan (limited to 20,000 requests per month). The date and time of each traffic observation are recorded in London local time at the time of data retrieval with a zoom level of 17, and~speed is reported in kilometres per hour (km/h). To~ensure temporal consistency across data sources, timestamps from all sources are standardised to Coordinated Universal Time (UTC). For~traffic data recorded in London local time, the~conversion to UTC accounts for the seasonal clock changes between Greenwich Mean Time (GMT) and British Summer Time (BST).

The raw traffic variables collected are current travel time (s), current speed (km/h), and~the free-flow travel time expected under ideal conditions (s). Direct traffic volume measurements (e.g., vehicle counts) are not available from the traffic data source used in this study. Instead, traffic conditions were characterised using speed and travel time. TomTom measures travel time and derives the corresponding speed based on the length of each road segment. Notably, the~current travel time is never lower than the free-flow travel time and, consequently, the~current speed is never higher than the free-flow speed. This behaviour suggests that traffic data may be capped when vehicle speeds exceed the road's speed~limit. 

From the raw traffic variables, traffic level is calculated as the ratio of current speed to free-flow speed, providing a relative measure of traffic conditions with respect to free-flow conditions. Traffic level and current travel time are subsequently used as the traffic-related predictors in the models. Traffic variables are averaged using all available observations within each hour (e.g., 09:00--09:59) and assigned to the corresponding hour-ending time\-stamp (e.g., 10:00) to match the temporal convention used for the atmospheric pollutant measurements.

Meteorological data are obtained from the OpenWeather platform~\cite{open_weather} by querying hourly historical records through the One Call 3.0 API, accessed through the “One Call by Call” subscription with up to 1000 API calls per day free of charge. The meteorological variables include temperature (K), sea-level atmospheric pressure (hPa), humidity (\%), dew point temperature (K), cloudiness (\%), wind direction (0--359$^\circ$), and~wind speed (m/s). For~both meteorological and traffic data sources, the~latitude and longitude coordinates of the target locations are used to retrieve the corresponding~observations.

To assess the suitability of the OpenWeather data, temperature and wind speed records are compared against observations from the target air quality monitoring stations, as~these are the only meteorological variables available from both data sources. Temperature shows a very strong agreement at both Camden Kerbside and Marylebone Road, with~Pearson correlation coefficients of 0.97 and fitted regression slopes close to unity (1.04 and 1.03, respectively). Wind speed also shows a positive association, although~with greater dispersion, with~Pearson correlation coefficients of 0.81 at Camden Kerbside and 0.74 at Marylebone Road, and~regression slopes of 1.04 and 0.89, respectively. These comparisons support the suitability of OpenWeather data for use as meteorological predictors in the models, while indicating that wind-speed estimates are subject to greater uncertainty than temperature. The~corresponding scatter plots and linear regression fits are provided in Figure~S1 of the Supplementary Material (see Appendix~\ref{App}).

Temporal variables are derived from the timestamps of the target station data and include the hour of the day (0--23), day of the week (0--6), and~month of the year (1--12). To preserve their cyclical nature, all temporal variables and wind direction are transformed into sine and cosine components.

For the neighbouring stations, unlike ref.~\cite{Samad2023}, which used the same monitoring stations as predictors for multiple target locations separated by distances ranging from 4 to 7 km, this study selects the two nearest traffic-type stations and the two nearest background-type stations independently for each target location from a set of 13 monitoring stations located within the same study area (Central London).

The nearest stations are identified by computing the Haversine distance using the implementation available in Scikit-learn~\cite{scikit-learn}, based on the latitude and longitude coordinates of each station and assuming an Earth radius of 6371 km. Table~\ref{tab:distance_matrix_between_stations} presents the target locations, the~selected neighbouring stations, and~the corresponding distances used in Scen~2 and Scen~3. Table~\ref{tab:Pollutants_Measured} lists the pollutants measured at each selected auxiliary~station.

\begin{table}[H]
\centering
\small
\caption{Pollutants Measured in the neighbouring stations considered in Scen~2 and Scen~3.}
\label{tab:Pollutants_Measured}
\newcolumntype{L}[1]{>{\raggedright\arraybackslash}p{#1}}
\setlength{\tabcolsep}{3pt}
    \begin{tabular}{L{5cm}L{8.5cm}}
        \toprule
        \textbf{Station} & \textbf{Pollutants Measured} \\
        \midrule
        Wandsworth--WA9 & \PMten, \NOtwo \\
        \midrule
        London Bloomsbury &
        \Othree, Nitric oxide (NO), \NOtwo, Nitrogen oxides as \NOtwo, \SOtwo, \CO, \PMten, \PMtwofive \\
        \midrule
        London N. Kensington &
        NO, \NOtwo, Nitrogen oxides as \NOtwo, \SOtwo, \CO, \PMten, \PMtwofive \\
        \midrule
        London Westminster &
        \Othree, NO, \NOtwo, Nitrogen oxides as \NOtwo, \PMtwofive \\
        \midrule
        Wandsworth--Lavender Hill & \NOtwo, \PMten \\
        \midrule
        Richmond Upon Thames--Castelnau & \NOtwo, \PMten \\
        \midrule
        RBKC Knightsbridge & \NOtwo \\
        \midrule
        Brent - ARK Franklin Primary Academy & \PMten and \NOtwo \\
        \midrule
        Camden High Street & \NOtwo \\
        \bottomrule
    \end{tabular}
\end{table}

Finally, observations from all data sources, including the neighbouring stations, are merged based on the target station and their corresponding UTC timestamps. As~part of data preprocessing, variables (columns) containing only missing values are removed. Negative values, which are considered invalid for specific variables, are treated as missing values (NaN). These files are stored for each target~station.

Correlation matrices between meteorological, temporal, and~traffic-related variables and the pollutant concentrations measured at each target station are calculated and provided in the Supplementary Material (Figures~S2--S6). Correlations among the traffic-related variables range from $-0.94$ to $-1.00$, depending on the station. 
The Spearman correlation coefficients of the traffic variables with the pollutants ranged from $-0.32$ to $0.53$.

The machine learning models explained in Section~\ref{sec:Machine-learning-workflow} are trained using a nested cross-validation framework~\cite{varma2006bias, cawley2010over} combined with the TimeSeriesSplit implementation from Scikit-learn~\cite{sklearn_timeseriessplit}. 
Nested cross-validation consists of an outer loop for model evaluation and an inner loop for hyperparameter optimisation. Unlike conventional cross-validation, where samples may be randomly assigned to folds, TimeSeriesSplit generates chronologically ordered training–test splits, preventing data leakage from future to past. The~first split contains approximately $\frac{2n}{k+1}$ observations, where $n$ is the total number of observations, and $k$ is the number of splits. Each subsequent split expands the training set by approximately $\frac{n}{k+1}$ observations, reflecting a realistic operational setting in which additional observations become available over time and can be incorporated into model retraining before future predictions are generated. Since the observations remain in chronological order, each split uses the earliest portion of the time series for model training and validation, whereas the immediately following chronological block is reserved for testing. The~test set always has a size of approximately $\frac{n}{k+1}$, ensuring that model performance is evaluated over test sets of comparable size across all~splits. 

The datasets used in each outer and inner fold are generated and stored prior to model training to ensure reproducibility. These datasets were generated separately for each station–pollutant pair. During~their generation, rows with missing target pollutant values were removed to avoid generating artificial target values. A~five-fold TimeSeriesSplit was applied to define the outer folds, and~a three-fold TimeSeriesSplit was applied within each outer training set to define the inner folds. Missing values corresponding to internal gaps of up to four consecutive observations are imputed using time-based linear interpolation.%, applied independently to the training and test sets of each outer fold to avoid data leakage. 
To avoid data leakage, imputation was performed independently within the training and validation sets of each inner fold and, subsequently, within the training and test sets of each outer fold.
The~same predefined outer and inner folds and corresponding timestamps were used to evaluate the tree-based models, Ridge regression, and~interpolation methods, ensuring that their performance metrics are calculated over identical observations.

At each target monitoring station, 4152 hourly observations were available prior to preprocessing. After~merging all data sources and applying the preprocessing procedures described in this section, the~number of hourly observations available before creating the outer and inner folds ranges from 3232 to 3912 per station–pollutant combination. \mbox{Tables~S1~and~S2} in the Supplementary Material summarise the number of hourly observations at each stage of the preprocessing pipeline for each station–pollutant combination.

\subsection{Feature~Scenarios}
\label{sec:feature_combination}

Six predictor scenarios are constructed to assess the contribution of traffic and spatial information.
Three base scenarios are adapted from~\cite{Samad2023}, we use Scen~1 to represent a situation in which no air quality monitoring network is available. Scen~2 represents a situation in which an air quality monitoring network is available, but~the number of stations is limited or only a few parameters are measured. In~this case, the~nearest background station is used because its measurements are representative of air quality over an area spanning several kilometres. Scen~3 represents a situation in which a denser monitoring network comprising both background and traffic monitoring stations is available. 
Following previous studies that compared models with and without traffic-related variables~\cite{DEVEER2025, Sulaimon2022}, each base scenario is evaluated both with and without traffic information, resulting in six predictor scenarios. The~predictor variables included in each scenario are:

\begin{itemize}
    \item Scen 1 (baseline): Temporal, meteorological, and~traffic-related variables at the target~location.
    
    \item Scen 2: Variables from Scen 1 plus the target pollutant measured at the nearest background~station. 
    
    \item Scen 3: Variables from Scen 2 plus all pollutants measured at the two nearest background stations and the two nearest traffic stations. 
\end{itemize}

\subsection{Machine Learning Workflow for Pollutant~Estimation}
\label{sec:Machine-learning-workflow}
Ensemble methods are machine learning techniques that combine multiple predictive models to generate a single, more robust, and~accurate prediction. These methods demonstrate strong performance across a wide range of applications, including air pollution prediction~\cite{Review_ensemble_air_pollution, ensemble_2024_PM10}. For~this reason, four tree-based ensemble algorithms are considered in this work: Random Forest (RF)~\cite{breiman2001randomforests}, Extra Trees (ET)~\cite{geurts2006extremely}, eXtreme Gradient Boosting (XGBoost)~\cite{ChenXGBoost}, and~Light Gradient Boosting Machine (LightGBM)~\cite{ke2017lightgbm}. These models are selected based on their strong predictive performance reported in previous benchmark studies and their recurrent identification among the best-performing algorithms across different pollutants and predictor configurations~\cite{virtual_monitoring_stations, Samad2023, Sulaimon2022, DEVEER2025, Du2025}.

The four tree-based machine learning models are trained using nested cross-validation with the folds generated by the TimeSeriesSplit procedure described previously. The~first four outer folds generated by TimeSeriesSplit are %were
used exclusively for model development. Within~each outer fold, a~three-fold TimeSeriesSplit was applied in the inner loop to optimise the hyperparameters of each candidate model. After~completing the first four outer folds, the~prediction algorithm was selected as the one achieving the lowest mean Root Mean Squared Error (RMSE) across the corresponding outer-fold test~sets. 

Finally, the~selected algorithm was trained using the fifth chronological split, where the training set comprised approximately five-sixths of the complete dataset. Hyperparameter optimisation was again performed within this training set using a three-fold TimeSeriesSplit, after~which the model was retrained on the complete training partition using the selected hyperparameter configuration. The~final model performance was then evaluated on the remaining one-sixth of the data, which had not been used at any stage of model selection or hyperparameter optimisation. Consequently, the~reported test results correspond to a completely independent chronological test set, thereby avoiding model-selection~bias. 

Hyperparameter optimisation was performed using Optuna v4.8.0~\cite{optuna_2019}, an~open-source optimisation framework, with~the default TPESampler and MedianPruner settings. The~explored hyperparameter search spaces are reported in Table~S3 of the Supplementary Material. A~total of 100 optimisation trials were conducted for each model, and~early stopping was implemented through Optuna's median pruning strategy to reduce computational cost. To~ensure reproducibility, all experiments were performed using a fixed random seed of 752 on the same computational environment, comprising Python 3.9.25 and 32 GB of~RAM. 

The models were trained independently for each of the six feature scenarios. Model performance was evaluated using the RMSE, Mean Absolute Error (MAE), and~coefficient of determination ($R^2$).

In addition to the tree-based machine learning models, Ridge Regression~\cite{Hoerl_Ridge} was included as a regularised linear regression baseline. This enables a direct comparison between linear and nonlinear regression models using the same predictor information. Ridge Regression uses the same predefined outer and inner folds for each station–pollutant pair. Since the preceding interpolation step only fills internal gaps, median imputation (SimpleImputer (strategy = ``median'')) is included in the Ridge pipeline to handle any missing values remaining at the edges of the time series. The~imputer is fitted exclusively on the training partition and then applied to the corresponding validation or test partition. 

Statistical comparisons between scenarios with and without traffic are performed on the independent chronological test set corresponding to the fifth split of the TimeSeriesSplit procedure. For~each station–pollutant combination, predictions from the compared scenarios are paired by timestamp. To~account for serial dependence among hourly observations, a~paired non-overlapping 24-h block bootstrap with 10,000 replicates is applied. Within~each bootstrap replicate, complete daily blocks are resampled with replacement, preserving the temporal dependence within each day and avoiding the assumption of independent hourly prediction errors. The~bootstrap procedure is used to estimate differences in RMSE, relative RMSE changes, 95\% confidence intervals, and~two-sided bootstrap \emph{p}-values. Because~multiple scenario comparisons are performed for each station–pollutant combination, the~resulting \emph{p}-values are adjusted using the Holm procedure to control the family-wise error rate. The~95\% confidence intervals are reported to quantify the uncertainty of the estimated RMSE changes, whereas statistical significance is determined using the Holm-adjusted \emph{p}-values. An~improvement resulting from the inclusion of traffic-related variables in model performance was considered statistically significant only when the relative RMSE change was positive and the Holm-adjusted \emph{p}-value was below 0.05.

Finally, SHAP (Shapley Additive exPlanations~\cite{Lundberg2017SHAP, lundberg2020local}) analysis is applied to all best-performing models to characterise the contribution of traffic-related variables to their predictions. Although~the SHAP beeswarm plots are presented for all best-performing models, the~interpretation of traffic-related feature importance focuses on cases in which the ablation analysis demonstrated incremental predictive value from traffic variables, as~indicated by a reduction in RMSE when traffic was included. SHAP importance reflects the contribution of variables to the predictive behaviour of the fitted model and should not be interpreted as evidence of direct causal effects~\cite{Janzing2020}.

\subsection{Spatial Interpolation~Baselines}

Kriging and Inverse Distance Weighting (IDW) are spatial interpolation methods widely used to estimate air pollutant concentrations at unsampled locations~\cite{Krigin_DTW_2024, GaussianProcesses2025, carroll2025pm25, mirza2026airquality, samal2023airquality, wadekar2026_krigin_IDW_approach}. In this study, both methods are employed as conventional spatial interpolation baselines for benchmarking the performance of the machine learning models, rather than as primary modelling approaches. The~interpolation is intentionally restricted to the same four neighbouring monitoring stations (the two nearest background stations and the two nearest traffic stations listed in Table~\ref{tab:distance_matrix_between_stations}) used for the corresponding target location. Station coordinates are projected from WGS84 (EPSG:4326) to the British National Grid (EPSG:27700), so that spatial distances are represented in metric~units.

To perform Ordinary Kriging, the~empirical variogram is estimated using measurements of the target pollutant from the target station and four neighbouring monitoring stations. Spherical, exponential, and~Gaussian variogram models were considered as alternative specifications. To~assess sensitivity to the choice of variogram while maintaining consistency with the ML model-selection procedure, the~three candidate variogram models were evaluated over the first four chronological outer folds. Within~each fold, the~variogram and its parameters were estimated exclusively from the corresponding training set and subsequently held fixed when interpolating concentrations at the target station in the outer-fold test set. The~model with the lowest mean RMSE across the four folds was selected. The~selected variogram model was then refitted using the training portion of the fifth chronological split and evaluated on its independent test set. Ordinary Kriging interpolation was implemented using PyKrige v1.7.3~\cite{pykrige2022}. The~final selected variogram model and its fitted parameters such as the partial sill, range, and~nugget are reported in Table~S4 of the Supplementary Material.

IDW estimates values as a weighted average of neighbouring observations, where the weights are inversely proportional to the Euclidean distance between locations raised to a power parameter $p$. In~this study, the~commonly used value $p=2$ is adopted, giving greater influence to nearby observations while reducing the contribution of more distant ones. This method assumes that nearby observations are more similar than distant ones and therefore exert a greater influence on the interpolated values~\cite{carroll2025pm25, fatima2022malaria}. For~each timestamp in the independent test set of the fifth chronological split, the~target pollutant concentration is estimated using simultaneous measurements of the same pollutant from the four neighbouring monitoring stations. Observations with incomplete measurements from these stations are excluded from the evaluation.

\subsection{Performance~Metrics}

Model performance is evaluated using the RMSE and MAE (both expressed in \umg), as~these error-based metrics directly quantify estimation accuracy in physical units and facilitate comparison with previous studies~\cite{Samad2023, Sulaimon2022}. The~coefficient $R^2$ is also computed on the test set as a supplementary performance~indicator.

The residuals given by $e_i$ in Equation~(\ref{eq:residuals}) are defined as the difference between the observed and predicted values for each test instance. The~distribution of the residuals is used to visualise the differences between models. Since a good model fit is generally associated with residuals centred around zero, residual analysis also provides an indication of model performance.
\begin{equation}
e_i = y_i - \hat{y_i}
\label{eq:residuals}
\end{equation}
    
\begin{itemize}
    \item RMSE is the square root of the average of the square residuals (see Equation~(\ref{eq:RMSE})). It is expressed in units of the dependent variable. It penalises larger errors more than smaller errors, since it squares the differences.
\begin{equation}
    \text{RMSE} = \sqrt{\frac{1}{n} \sum_{i=1}^{n} (y_i - \hat{y_i})^2}
    \label{eq:RMSE}
    \end{equation}
    
    \item MAE is a metric that calculates the average magnitude of the absolute difference between the predicted and the observed values (see Equation~(\ref{eq:MAE})). It is also expressed in units of the dependent variable. MAE is more robust to outliers in the data as it does not penalise larger errors more than smaller errors, since it does not square the~differences.
\begin{equation}
    \text{MAE} = \frac{1}{n} \sum_{i=1}^{n} |y_i - \hat{y_i}|
    \label{eq:MAE}
    \end{equation}

\end{itemize}

The percentage improvement is also calculated for pairwise comparisons between scenarios, where A denotes the reference (baseline) scenario and B the alternative scenario. Since B always includes a larger set of predictors than A, positive values indicate an improvement in predictive performance (i.e., a~lower RMSE) achieved by incorporating additional predictors, whereas negative values indicate a degradation relative to the baseline~scenario.
\begin{equation}
\Delta\text{RMSE} (\%) = 
\frac{\text{RMSE}_{A} - \text{RMSE}_{B}}
{\text{RMSE}_{A}} \times 100
\label{eq:percentage_improvement}
\end{equation}

Conversely, the~percentage degradation in predictive performance (i.e., a~higher RMSE) is calculated using Equation~(\ref{eq:degradation}).
\begin{equation}
\label{eq:degradation}
\text{Degradation}(\%) =
\frac{\mathrm{RMSE}_B - \mathrm{RMSE}_A}
{\mathrm{RMSE}_A}
\times 100
\end{equation}

\section{Results}
\label{sec:Results}

For each station–pollutant combination, the~algorithm with the lowest mean RMSE across the first four outer folds is selected as the best-performing model and subsequently retrained and hyperparameter-tuned using the fifth chronological split. The~performance metrics presented in this section correspond to that model evaluated on the independent chronological test set.

\begin{figure}[H]
\centering
    \includegraphics[width=.8\linewidth]{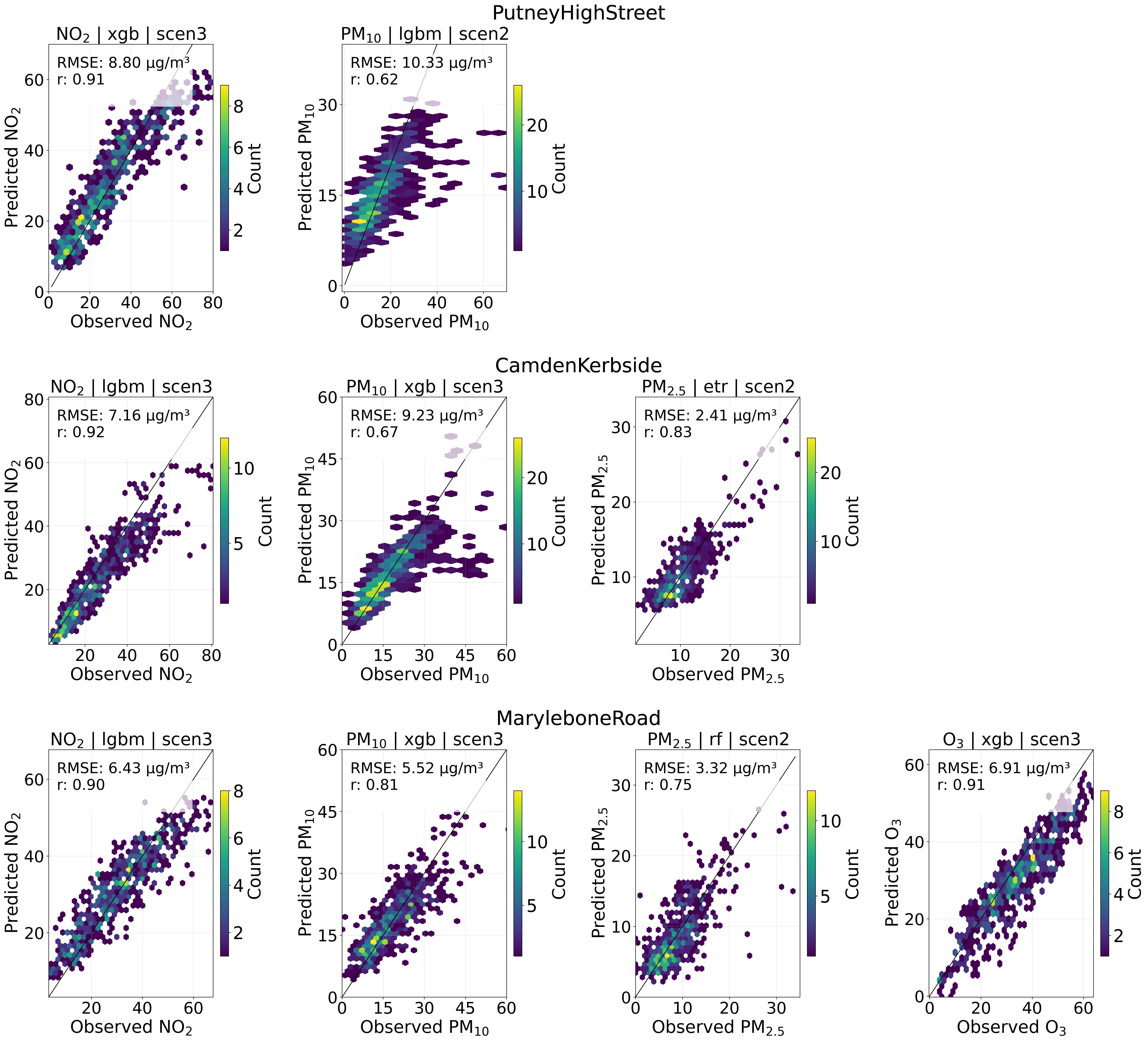}
    \caption{Observed vs predicted concentrations for the best-performing model (scenario and model) per station--pollutant pair. RMSE and the Pearson correlation coefficient ($r$) are shown in the upper-left corner of each~panel.} 
    \label{fig:Obs_vs_pred}
\end{figure}

Figure~\ref{fig:Obs_vs_pred} shows the observed and predicted pollutant concentrations obtained with the best-performing model for each station–pollutant combination, together with the 1:1~line representing perfect agreement between observations and predictions. The title of each panel indicates the machine learning model name and predictor scenario. 
The upper-left corner of each panel reports the RMSE and the Pearson correlation coefficient ($r$).

Figure~\ref{fig:Obs_vs_pred} shows that the agreement between observed and predicted concentrations varies across pollutants and monitoring sites. While the models generally capture the relationship between observed and predicted concentrations for \NOtwo and \Othree, deviations from the 1:1 line become more evident at concentration extremes for \PMten and \PMtwofive. In particular, some models tend to overestimate lower concentrations and underestimate higher concentrations. The~Pearson correlation coefficient between observed and predicted concentrations is approximately 0.9 for \NOtwo and \Othree, indicating a strong linear association between predictions and observations. For~\PMten and \PMtwofive, the correlation is more variable across monitoring sites, with~coefficients ranging from approximately 0.62 to 0.83.

To further assess whether the models reproduce the temporal variability of pollutant concentrations, observed and predicted concentrations are additionally presented as time series in Figures~S7--S10 of the Supplementary Material. For~\NOtwo, the time-series comparison shows that, at~most sites, predicted and observed concentrations generally exhibit similar temporal patterns, with~increases and decreases occurring broadly in phase. In~contrast, the~time series for \Othree, \PMten, and~\PMtwofive show larger discrepancies in magnitude between observed and predicted concentrations, particularly at concentration peaks.
This complementary analysis therefore indicates that the models can capture relevant temporal dynamics even when individual concentration extremes are not reproduced with the same~accuracy.

\begin{figure}[H]
  \centering
  \includegraphics[width=0.8\linewidth]{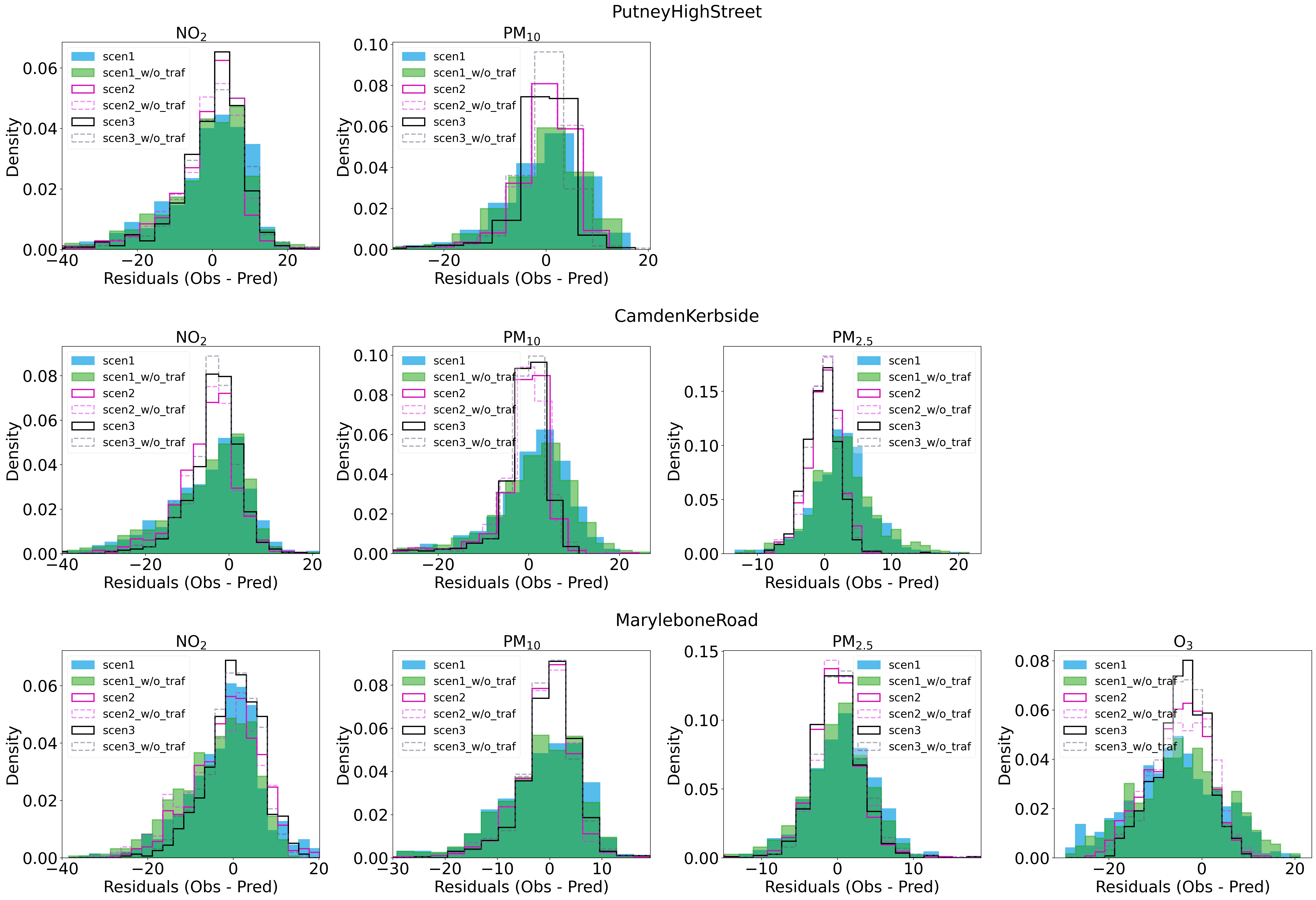}
    \caption{Distribution of residuals for each scenario (with and without traffic) of the best-performing model. Distributions of Scen 1 with and without traffic are highlighted in blue and light green, respectively, for~improved visual distinction (creating dark green where the two overlap).}
    \label{fig:Residuals_distribution}
\end{figure}

Figure~\ref{fig:Residuals_distribution} shows the distri\-bution of the resi\-duals of the best-per\-forming models, for~each station–pollutant–scenario combination. All scenarios (with and without traffic) are displayed to facilitate direct comparison. 
Residuals are generally centred around zero across most station--pollutant pairs, with~Scen~2 and 3 having larger peaks at zero and a narrower distribution than Scen~1. In~the case of \NOtwo, there is a negative skew that is more apparent in Scen~1, suggesting that in this case the model under-predicts extreme values. This improves for Scen~2 and 3, which suggests the extra information improves model~performance.

Table~\ref{tab:rmse_improvement} details the RMSE values of the best-performing model for each pollutant–station–scenario combination. 
For completeness, Table~S5 in the Supplementary Material reports the mean and standard deviation of the RMSE, MAE, and~$R^2$ across the first four outer folds of the nested cross-validation procedure for the selected model corresponding to each station–pollutant–scenario combination.

\begin{table}[H]
\centering
\caption{RMSE values per scenario and percentage improvement across scenarios ($\Delta$RMSE(\%) defined in Equation~(\ref{eq:percentage_improvement})).} 
\label{tab:rmse_improvement}
\setlength{\tabcolsep}{3pt}
\renewcommand{\arraystretch}{1}
\resizebox{\textwidth}{!}{%
    \begin{tabular}{cllcccccccccccc}
        \toprule
        \textbf{Case} & \textbf{Pollutant} & \textbf{Station} & \textbf{S1-noT} & \textbf{S1} & \textbf{S2-noT} & \textbf{S2} & \textbf{S3-noT} & \textbf{S3} & \boldmath{$\Delta$}\textbf{S1} & \boldmath{$\Delta$}\textbf{S2} & \boldmath{$\Delta$}\textbf{S3} & \boldmath{$\Delta$}\textbf{S2/S1} & \boldmath{$\Delta$}\textbf{S3/S1} & \boldmath{$\Delta$}\textbf{S3/S2} \\
        \midrule
        1 & \NOtwo   & Camden Kerbside              & 11.43 & 11.27 &  8.94 &  8.86 &  7.50 &  7.16 & 11.46 &  0.91 &  4.63 & 21.34 & 36.48 & 19.25 \\
        2 & \NOtwo  & Marylebone Road        &  9.73 &  8.72 &  8.77 &  7.96 &  6.80 &  6.43 & 10.38 &  9.28 &  5.44 &  8.77 & 26.24 & 19.15 \\
        3 & \NOtwo  & Putney High Street & 11.66 & 11.52 &  9.81 &  9.37 &  9.45 &  8.80 & 11.21 &  4.53 &  6.86 & 18.68 & 23.61 &  6.06 \\
        4 & \Othree  & Marylebone Road        & 10.58 & 11.20 &  8.04 &  8.21 &  7.22 &  6.91 & $-$5.79 & $-$2.12 &  4.21 & 26.69 & 38.25 & 15.77 \\
        5 & \PMten & Camden Kerbside              & 11.34 & 10.53 &  9.72 &  9.70 &  9.31 &  9.23 &  7.14 &  0.14 &  0.88 &  7.88 & 12.38 &  4.88 \\
        6 & \PMten & Marylebone Road        &  8.36 &  8.43 &  5.85 &  5.84 &  5.57 &  5.52 &  0.90 &  0.26 &  0.91 & 30.82 & 34.53 &  5.37 \\
        7 & \PMten & Putney High Street & 12.09 & 12.05 & 10.35 & 10.33 & 10.44 & 10.51 &  0.33 &  0.13 & $-$0.68 & 14.24 & 12.77 & $-$1.71 \\
        8 & \PMtwofive %MDPI: we revised PM25 to PM2.5, please confirm. The following highlights are the same.
 & Camden Kerbside              &  5.49 &  4.93 &  2.43 &  2.41 &  2.42 &  2.54 & 10.19 &  0.63 & $-$5.14 & 51.04 & 48.45 & $-$5.28 \\
        9 & \PMtwofive & Marylebone Road        &  4.38 &  4.54 &  3.37 &  3.32 &  3.46 &  3.38 & $-$3.64 &  1.32 &  2.30 & 26.76 & 25.45 & $-$1.79 \\
        \bottomrule
    \end{tabular}%
    }

\noindent{\footnotesize{Note: S1, S2 and S3 denote scenarios 1, 2 and 3; noT denotes the scenario version without traffic variables. In~$\Delta$S(1,2,3), Sx with traffic is always the alternative scenario, and~Sx-noT is the reference one; positive values therefore indicate improvement when traffic variables are added, whereas negative values indicate degradation. For~$\Delta$Sx/Sy, both Sx and Sy are traffic scenarios, where the first one (Sx) is the alternative scenario and the second one is the reference scenario.}}

\end{table}

Table~\ref{tab:rmse_improvement} also reports the percentage improvement ($\Delta$RMSE) (given by Equation~(\ref{eq:percentage_improvement})) for pairwise comparisons between scenarios. 
For pairwise comparisons between scenarios with and without traffic predictors ($\Delta$Sx, where x = 1, 2, or~3), the~scenario including traffic-related variables (Sx) is always treated as the alternative scenario, whereas the corresponding scenario without traffic (Sx-noT) is used as the reference scenario in Equation~(\ref{eq:percentage_improvement}). Consequently, positive values indicate an improvement in predictive performance when traffic variables are included, while negative values indicate a~degradation.

To calculate the improve\-ment between sce\-narios that include traffic variables ($\Delta$S2/S1, $\Delta$S3/S1 and $\Delta$S3/S2),
the scenario with the larger predictor set is treated as the alternative scenario, while the simpler configuration is used as the reference scenario in Equation~(\ref{eq:percentage_improvement}). Therefore, positive values indicate an improvement in predictive performance resulting from the inclusion of additional predictor~variables. 

Across the majority of the evaluated scenarios, incorporating traffic-related variables reduces the RMSE. 
\NOtwo achieves a consistent positive improvement across the three stations in all the scenarios by adding traffic-related variables. 
However, for~\Othree, \PMten and \PMtwofive, there are a few specific cases in which adding traffic increases the RMSE; for example, for~\PMten at Putney High Street and \PMtwofive at Camden Kerbside under Scen~1 and Scen~2, the~traffic contributes to improving the performance of the model, while under Scen~3 it decreases the performance. When predicting \Othree in Scen~1 and Scen~2, adding traffic variables increases the RMSE, but~in Scen~3 there is a decrease. The~results suggest that the improvement in performance when adding traffic depends largely on the station and~pollutant.

\begin{figure}[H]
  \centering
  \includegraphics[width=.5\linewidth]{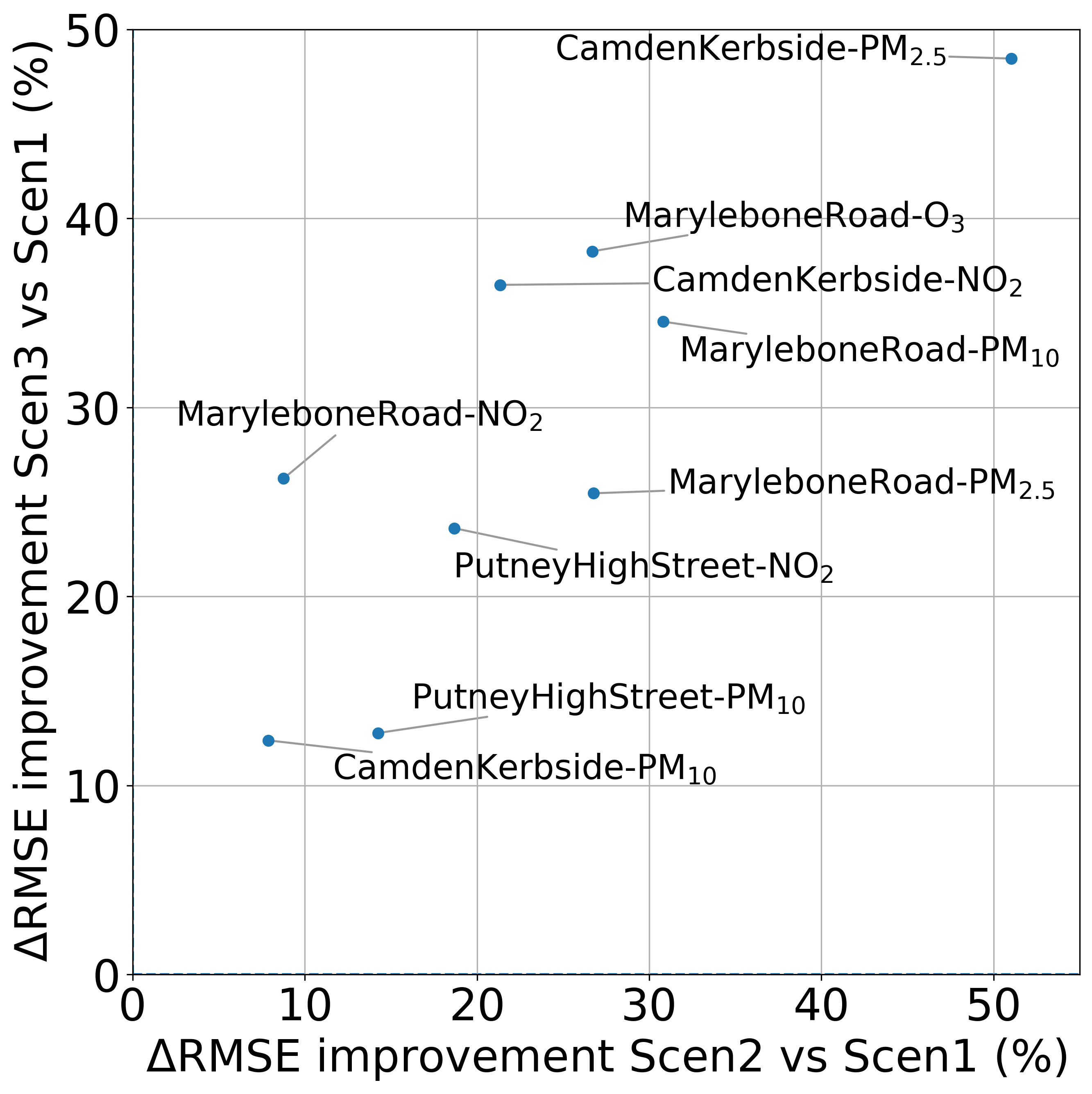}
  \caption{Percentage improvement in RMSE (calculated with Equation~(\ref{eq:percentage_improvement})) obtained by incorporating additional spatial pollutant information compared with using the simplest predictor set that includes traffic-related variables (Scen~1). The~x-axis shows the percentage improvement obtained by adding observations from the nearest auxiliary monitoring station (Scen~2 vs. Scen~1), while the y-axis shows the improvement achieved by incorporating multi-station information (Scen~3 vs. Scen~1). Each point corresponds to a station–pollutant combination.
    }
    \label{fig:performance_gain}
\end{figure}

Figure~\ref{fig:performance_gain} provides a graphical representation of the percentage RMSE improvements (Equation~(\ref{eq:percentage_improvement})) reported in Table~\ref{tab:rmse_improvement}. The~figure compares the improvements obtained by incorporating additional pollutant information from neighbouring stations through Scen~2 (x-axis) and Scen~3 (y-axis), relative to Scen~1, for~the best-performing model at each station--pollutant~combination.

Across all scenarios including traffic-related variables, adding the target pollutant measured at the nearest background monitoring station to the baseline predictor set (i.e., moving from Scen~1 to Scen~2) improves predictive performance for all station–pollutant combinations. Similarly, improvements are observed for all station--pollutant combinations when comparing the scenarios that include pollutant measurements from four neighbouring stations with the baseline (i.e., from~Scen~1 to Scen~3). 

In the case of \PMtwofive at Camden Kerbside, going from Scen~1 to Scen~2 has a large improvement (51.04\%), but~going from Scen~1 to Scen~3 has a smaller improvement (48.45\%), which could suggest that including pollutant measurements from the neighbouring stations does not provide further predictive benefit in this case and may introduce redundant~information.

Regarding the percentage improvement in RMSE from Scen~2 to Scen~3, the~improvement is relatively small (less than 6\%) for \NOtwo at Putney High Street and for \PMten at all sites. Performance even decreases for \PMtwofive and \PMten at Putney High Street. 
This relatively small or negative improvement suggests that Scen~2 may provide a reasonable trade-off between predictive performance and data requirements across pollutants and sites, particularly when observations from multiple monitoring stations are unavailable. 
In contrast, for~\Othree and \NOtwo at the remaining stations, the~inclusion of data from the four neighbouring stations leads to a larger improvement in predictive performance (greater than 15\%).

\subsection{Analysis of Predictor~Contributions}

\begin{figure}[H]
  \centering
  \includegraphics[width=0.89\linewidth]{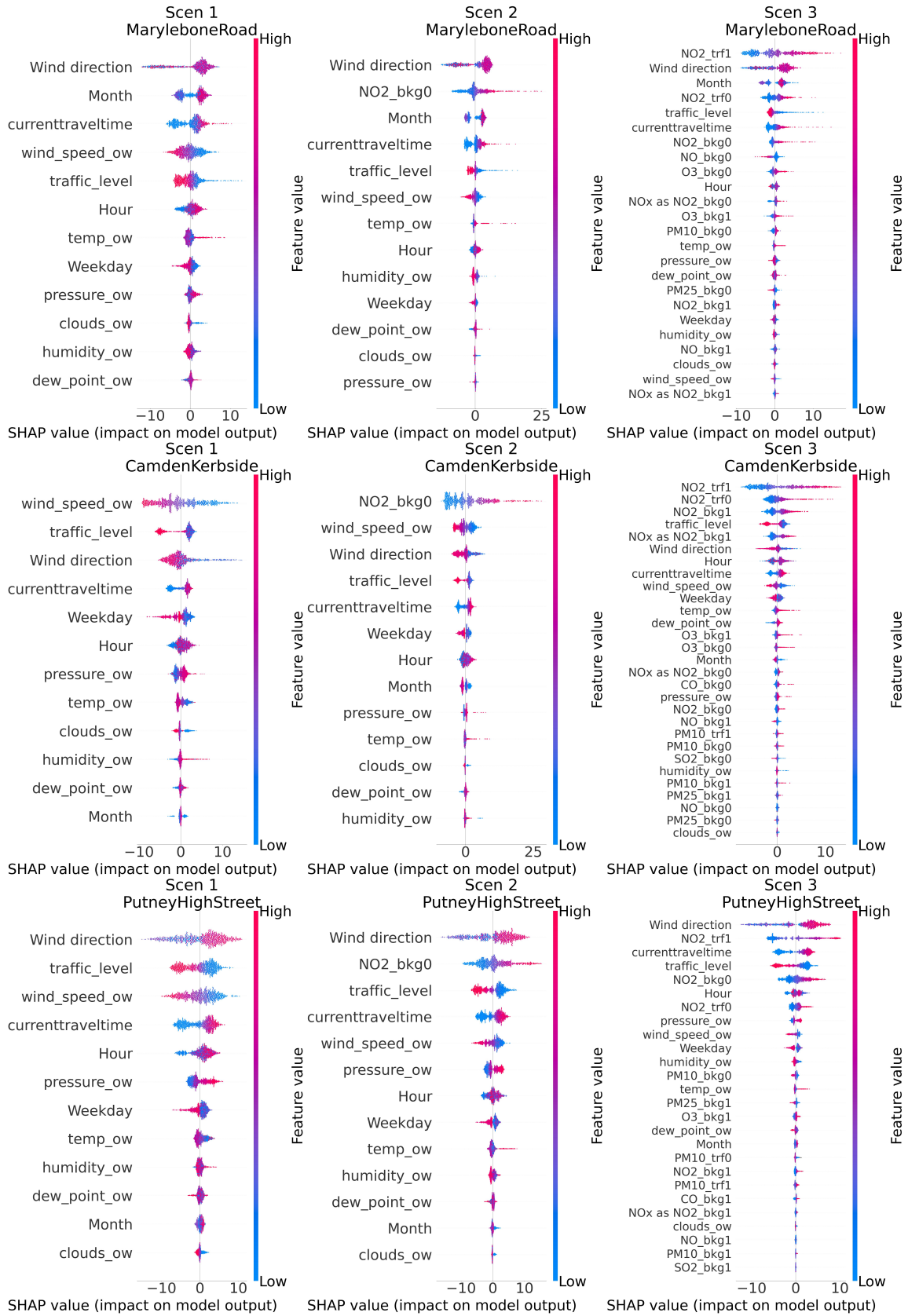}
  \caption{SHAP beeswarm plots of the best-performing models for \NOtwo predictions across stations and scenarios. Each point represents an observation, and~its colour indicates the feature value, ranging from low (blue) to high (red). Features are ranked according to their global SHAP importance, from~most important (top) to least important (bottom). The~horizontal spread indicates the magnitude of each feature's impact on the predicted pollutant concentration.
  }
  \label{fig:SHAP_NO2}
\end{figure}

\begin{figure}[H]
  \centering
  \includegraphics[width=0.89\linewidth]{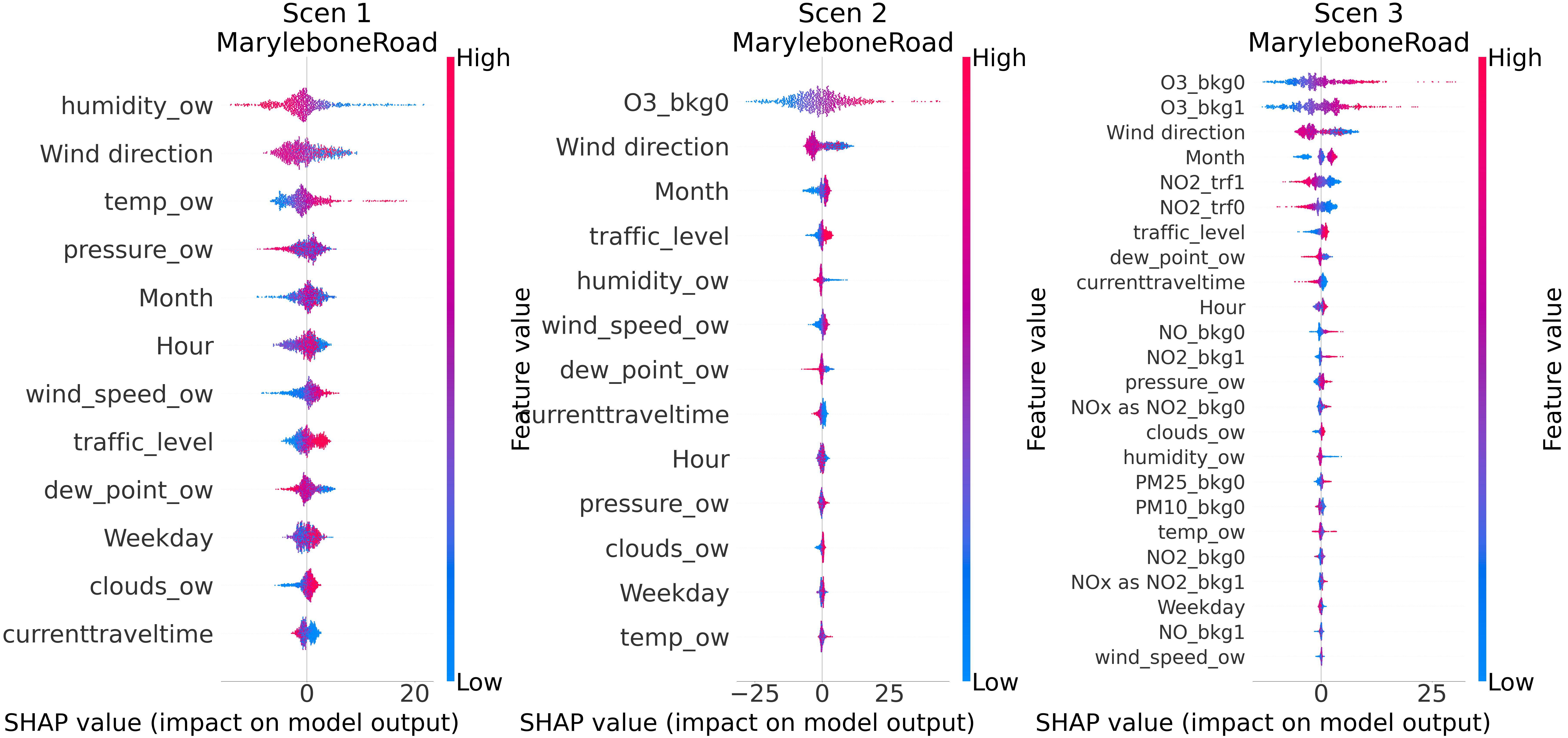}
  \caption{SHAP beeswarm plots of the best-performing models for \Othree predictions across stations and~scenarios.}
  \label{fig:SHAP_O3}
\end{figure}

\begin{figure}[H]
  \centering
  \includegraphics[width=0.89\linewidth]{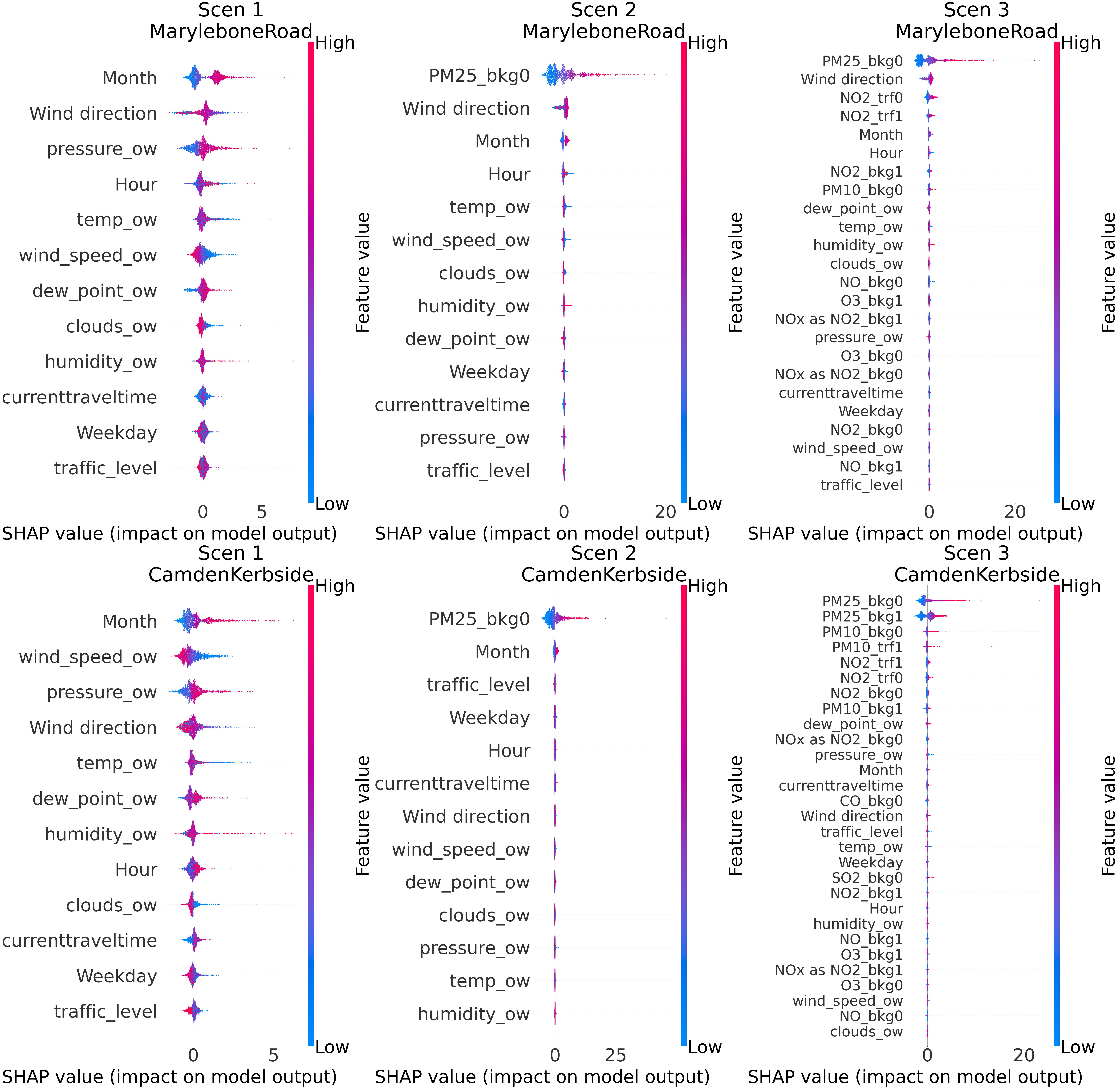}
  \caption{SHAP beeswarm plots of the best-performing models for \PMtwofive predictions across stations and~scenarios.}
  \label{fig:SHAP_PM25}
\end{figure}

To visualise the factors driving the model predictions when traffic information is included, Figures~\ref{fig:SHAP_NO2}--\ref{fig:SHAP_PM25} show SHAP beeswarm plots for the best-performing model in each station--pollutant-scenario combination, considering the scenario version with traffic variables. For the SHAP beeswarm plots, the~sine and cosine components of each cyclical temporal variable are represented as a single feature by summing their SHAP contributions for each observation. Accordingly, \textit{hour sin} and \textit{hour cos}, \textit{weekday sin} and \textit{weekday cos}, and~\textit{month sin} and \textit{month cos} are displayed as \textit{Hour}, \textit{Weekday}, and~\textit{Month}, respectively.

SHAP values quantify the contribution of each feature to the model predictions. In~these plots, features are ranked according to their global SHAP importance, from~most important (top) to least important (bottom). Each point represents a single observation. The~horizontal axis shows the SHAP value, indicating the contribution of a feature to the model output relative to the model baseline prediction. Positive SHAP values indicate that the feature increases the predicted pollutant concentration, whereas negative values indicate that it decreases the predicted concentration. Features with greater importance typically exhibit a wider horizontal spread of SHAP values. The~vertical spread reflects the density of observations with similar SHAP values and is used to reduce point overlap. Point colours represent the input feature values, ranging from low (blue) to high (red).

Table~S6 of the Supplementary Material reports the mean percentage contribution of each feature group (temporal, meteorological, traffic-related, and~auxiliary-station variables), which is quantified by aggregating the mean absolute SHAP values within each group. As~expected, the~overall pollutant predictions are strongly influenced by weather conditions, with~wind direction consistently ranked among the predictors with the largest SHAP contributions across almost all station–pollutant–scenario combinations. Temporal variables also made substantial contributions to the model predictions, particularly in Scenarios 1 and 2, where \textit{month} consistently ranked among the predictors with the largest SHAP contributions. This prominence suggests that recurrent temporal patterns provide valuable predictive information to the~models. 

Based on the ablation analysis used to identify cases in which traffic-related variables provided incremental predictive value (summarised in Table~\ref{tab:rmse_improvement}), \NOtwo~showed the most consistent benefit from the inclusion of traffic information, with~RMSE reductions across the three scenarios. The~subsequent SHAP analysis (Figure~\ref{fig:SHAP_NO2}) is used to examine the contribution of individual predictors within these fitted models. Traffic-related features consistently rank among the six most important predictors across the analysed scenarios, while meteorological variables are also among the main contributors.

At Camden Road and Putney High Street, traffic variables rank higher than temporal variables, and~at all stations traffic variables rank lower than nearest station pollutants (Table~S5 of the Supplementary Material). Traffic variables at Putney High Street have the highest percentage contribution in Scen~1 at 26.26\%, while Marylebone Road has the lowest at 11.88\%. These results indicate the contribution depends largely on the monitoring site, which could be due to many different factors such as road topology, pollutant distributions and traffic~profiles.  

When predicting \NOtwo, under~Scen~3, traffic-related variables across the three stations are frequently ranked above predictors derived from pollutants. The~only pollutant predictor variable ranked higher than traffic is \NOtwo. This indicates that, within~these fitted models, traffic information contributes to the predictions at a level comparable to, and~in some cases greater than, measurements of other pollutants from neighbouring monitoring stations, including \PMten, \PMtwofive, NO, CO, \Othree, and, in~some cases, NO$_x$ as \NOtwo and \SOtwo.

For \Othree, the ablation analysis shows an improvement from traffic-related variables only under Scen~3. Within~this model, the~SHAP analysis (Figure~\ref{fig:SHAP_O3}) ranks traffic-related variables above several pollutant measurements from neighbouring monitoring stations, including \PMten, \PMtwofive, NO, and~NO$_x$ as \NOtwo, after~the \NOtwo~measurements from the two nearest traffic stations had been considered. Thus, while the ablation analysis establishes the observed predictive benefit of including traffic under this scenario, SHAP provides complementary information on the relative contribution of traffic variables within the fitted~model. 

\begin{figure}[H]
  \centering
  \includegraphics[width=0.89\linewidth]{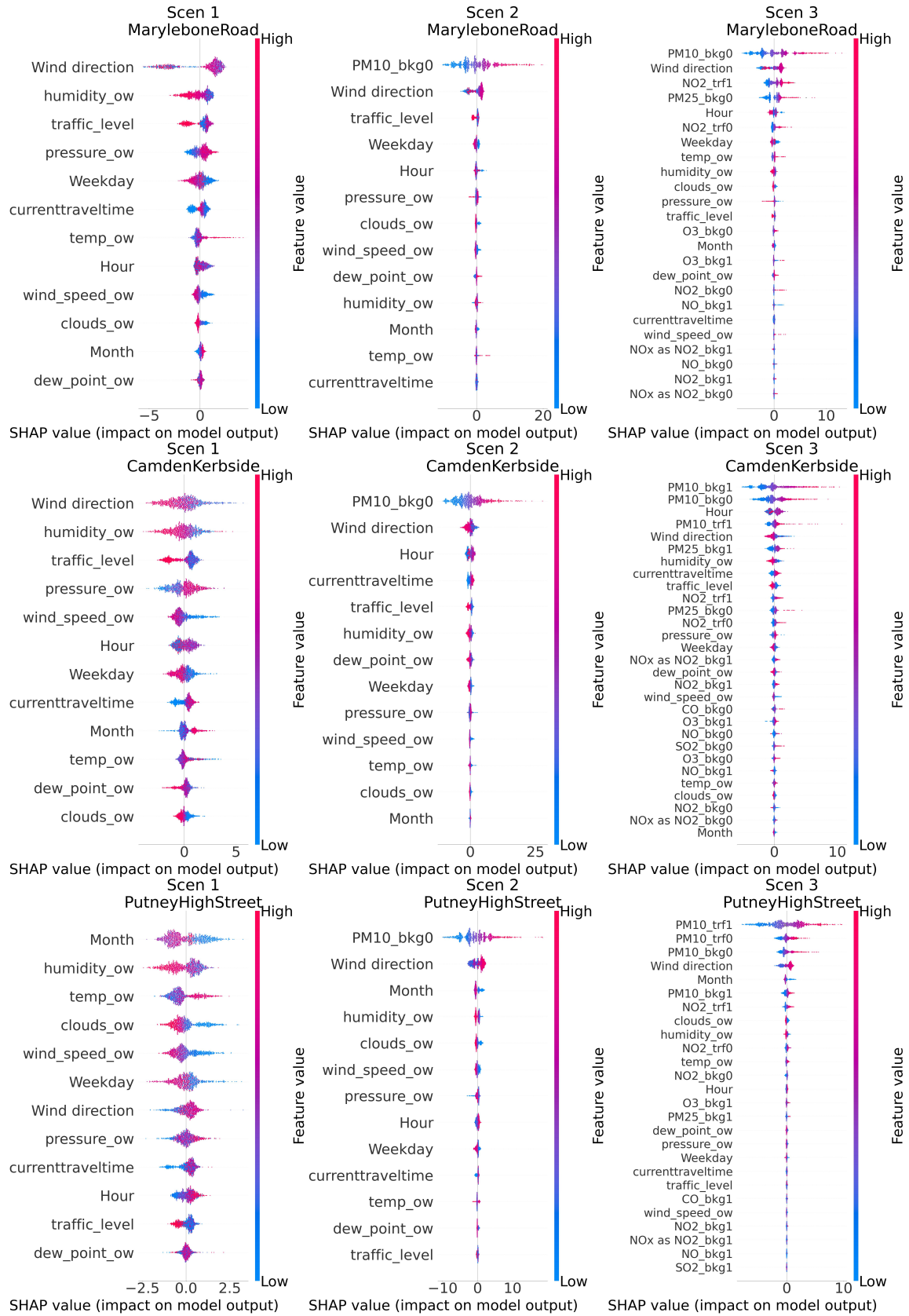}
  \caption{SHAP beeswarm plots of the best-performing models for $PM_{10}$ predictions across stations and~scenarios.}
  \label{fig:SHAP_PM10}
\end{figure}

For \PMtwofive and \PMten (Figure~\ref{fig:SHAP_PM25} and Figure~\ref{fig:SHAP_PM10}, respectively), even in cases where the ablation analysis shows an RMSE reduction after incorporating traffic-related variables, their SHAP contributions are generally small and traffic features tend to rank near the lower end of the feature-importance list. This indicates that, although~traffic information provides some incremental predictive benefit in these cases, the~fitted models rely on it considerably less than on other predictors.

\subsection{Comparison with Spatial Interpolation and Linear Regression~Baselines}
\label{sec:interpolation}

To further contextualise the performance of the tree-based machine learning models, their predictions are compared with those obtained using a linear regression model (Ridge regression) and two spatial interpolation methods (IDW and OK).

Tables~\ref{tab:ML_vs_Ridge_NO2_complete}--\ref{tab:ML_vs_Ridge_PM25_complete} present the performance metrics of the best-performing tree-based model and the Ridge regression model on the independent test set. Across the four pollutants, the~tree-based models generally outperform Ridge regression in terms of RMSE, MAE, and~$R^2$ across almost all stations and scenarios. In~some cases, Ridge regression yields negative $R^2$ values, indicating that its predictions perform worse than simply using the mean observed concentration as a predictor.

\begin{figure}[H]
  \centering
  \includegraphics[width=0.99\linewidth]{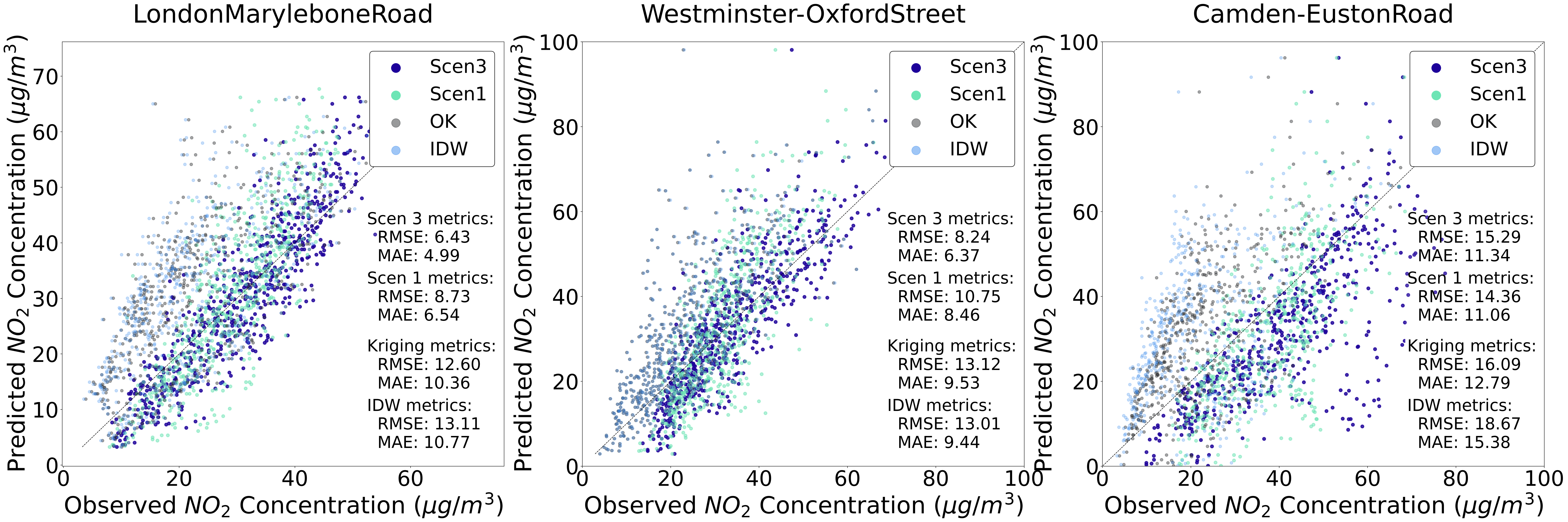}
  \caption{Observed versus predicted \NOtwo concentrations for each station, comparing machine learning models (Scen~1 and Scen~3) with spatial interpolation approaches (Ordinary Kriging and IDW). Interpolation was performed using measurements from four nearby stations of the same pollutant, providing a spatial baseline for~comparison.}
  \label{fig:interpolation}
\end{figure}

Both interpolation methods (OK and IDW) are used to estimate only \NOtwo, as~it is the only pollutant simultaneously measured at all four neighbouring stations. Figure~\ref{fig:interpolation} displays the observed versus predicted \NOtwo concentrations for each station, comparing the interpolation methods and the best-performing model per station–pollutant implemented with Scen~1 and Scen~3 (both with traffic variables). Figure~\ref{fig:interpolation} also reports the RMSE (\umg) and MAE (\umg) performance metrics. 

The predictive performance of both interpolation methods is inferior to that of the machine learning models, even when considering the simplest feature configuration (Scen~1), which only relies on local traffic, temporal and meteorological variables without incorporating information from neighbouring stations. This difference is reflected in both the dispersion of the predicted values and the reported error metrics, failing to capture the full variability of the observed concentrations, particularly at higher values.

%\unskip

%\unskip

%\vspace{-6pt}

%------------------------
\begin{table}[H]
\centering
\caption{\NOtwo model performance: RMSE, MAE, and $R^2$ for each station and scenario for the best-performing tree-based model and the Ridge regression model.
}
\label{tab:ML_vs_Ridge_NO2_complete}
\small

\begin{tabular}{lllcccccc}
\toprule

\textbf{Station} & \textbf{Scenario} & \textbf{Best ML} &
\multicolumn{3}{c}{\textbf{Best ML}} &
\multicolumn{3}{c}{\textbf{Ridge}} \\

\cmidrule{4-9}

 & & \textbf{Model} & \textbf{RMSE} & \textbf{MAE} & \boldmath{$R^2$} & \textbf{RMSE} & \textbf{MAE} & \boldmath{$R^2$} \\

\midrule
\multirow{6}{*}{Marylebone Road} & Scen1-noT & LGBM & 9.73 & 7.36 & 0.53 & 11.43 & 9.15 & 0.35 \\
 & Scen1 & XGB & 8.72 & 6.54 & 0.62 & 11.74 & 9.41 & 0.32 \\
 & Scen2-noT & ETR & 8.77 & 6.82 & 0.62 & 8.41 & 6.81 & 0.65 \\
 & Scen2 & XGB & 7.96 & 6.17 & 0.69 & 9.14 & 7.32 & 0.59 \\
 & Scen3-noT & LGBM & 6.80 & 5.21 & 0.77 & 7.16 & 5.77 & 0.75 \\
 & Scen3 & LGBM & 6.43 & 4.99 & 0.80 & 7.77 & 6.25 & 0.70 \\
\midrule
\multirow{6}{*}{Camden Kerbside} & Scen1-noT & XGB & 11.43 & 8.18 & 0.47 & 14.20 & 10.28 & 0.18 \\
 & Scen1 & XGB & 11.27 & 8.12 & 0.48 & 13.70 & 9.83 & 0.23 \\
 & Scen2-noT & LGBM & 8.94 & 6.59 & 0.67 & 10.61 & 8.05 & 0.54 \\
 & Scen2 & XGB & 8.86 & 6.47 & 0.68 & 10.45 & 7.83 & 0.56 \\
 & Scen3-noT & LGBM & 7.50 & 5.36 & 0.77 & 8.60 & 6.48 & 0.70 \\
 & Scen3 & LGBM & 7.16 & 5.12 & 0.79 & 8.50 & 6.39 & 0.70 \\
\midrule
\multirow{6}{*}{Putney High Street} & Scen1-noT & LGBM & 11.66 & 8.45 & 0.64 & 13.51 & 10.24 & 0.52 \\
 & Scen1 & XGB & 11.52 & 8.49 & 0.65 & 12.17 & 9.02 & 0.61 \\
 & Scen2-noT & LGBM & 9.81 & 6.99 & 0.74 & 10.02 & 7.49 & 0.73 \\
 & Scen2 & LGBM & 9.37 & 6.61 & 0.77 & 9.38 & 6.86 & 0.77 \\
 & Scen3-noT & LGBM & 9.45 & 6.94 & 0.76 & 8.87 & 6.66 & 0.79 \\
 & Scen3 & XGB & 8.80 & 6.37 & 0.80 & 8.31 & 6.17 & 0.82 \\
\midrule
\multirow{6}{*}{Oxford Street} & Scen1-noT & XGB & 11.42 & 8.88 & 0.51 & 13.38 & 10.96 & 0.32 \\
 & Scen1 & XGB & 10.74 & 8.45 & 0.56 & 12.48 & 10.37 & 0.41 \\
 & Scen2-noT & LGBM & 9.88 & 7.41 & 0.63 & 10.41 & 8.21 & 0.59 \\
 & Scen2 & LGBM & 9.01 & 6.86 & 0.69 & 9.36 & 7.52 & 0.67 \\
 & Scen3-noT & XGB & 9.30 & 6.89 & 0.67 & 9.55 & 6.91 & 0.66 \\
 & Scen3 & XGB & 8.24 & 6.37 & 0.74 & 9.31 & 7.34 & 0.67 \\
\midrule
\multirow{6}{*}{Euston Road} & Scen1-noT & LGBM & 12.84 & 10.00 & 0.46 & 13.09 & 9.91 & 0.44 \\
 & Scen1 & ETR & 14.33 & 11.03 & 0.33 & 14.39 & 10.83 & 0.32 \\
 & Scen2-noT & RF & 13.11 & 10.30 & 0.44 & 12.10 & 8.98 & 0.52 \\
 & Scen2 & XGB & 13.20 & 10.11 & 0.43 & 13.85 & 10.05 & 0.37 \\
 & Scen3-noT & RF & 12.98 & 10.41 & 0.45 & 10.03 & 7.42 & 0.67 \\
 & Scen3 & LGBM & 15.28 & 11.33 & 0.24 & 12.62 & 9.31 & 0.48 \\
\bottomrule
\end{tabular}
\end{table}
\unskip

%------------------------
\begin{table}[H]
\centering
\caption{\Othree model performance: RMSE, MAE, and $R^2$ for each station and scenario for the best-performing tree-based model and the Ridge regression model.
}
\label{tab:ML_vs_Ridge_O3_complete}
\small
\begin{tabular}{lllcccccc}
\toprule

\textbf{Station} & \textbf{Scenario} & \textbf{Best ML }&
\multicolumn{3}{c}{\textbf{Best ML}} &
\multicolumn{3}{c}{\textbf{Ridge}} \\
\cmidrule{4-9}
 & & \textbf{Model} & \textbf{RMSE} & \textbf{MAE} & \boldmath{$R^2$} & \textbf{RMSE} & \textbf{MAE} & \boldmath{$R^2$} \\

\midrule
\multirow{6}{*}{Marylebone Road} & Scen1-noT & LGBM & 10.58 & 8.47 & 0.38 & 15.49 & 12.82 & $-$0.34 \\
 & Scen1 & LGBM & 11.20 & 9.05 & 0.30 & 15.93 & 13.31 & $-$0.42 \\
 & Scen2-noT & LGBM & 8.04 & 6.32 & 0.64 & 8.55 & 6.82 & 0.59 \\
 & Scen2 & LGBM & 8.21 & 6.41 & 0.62 & 8.89 & 7.01 & 0.56 \\
 & Scen3-noT & LGBM & 7.22 & 5.66 & 0.71 & 7.20 & 5.94 & 0.71 \\
 & Scen3 & XGB & 6.91 & 5.42 & 0.73 & 7.80 & 6.38 & 0.66 \\
\bottomrule
\end{tabular}

\end{table}
\unskip

%------------------------
\begin{table}[H]
\centering
\caption{\PMten model performance: RMSE, MAE, and $R^2$ for each station and scenario for the best-performing tree-based model and the Ridge regression model.}
\label{tab:ML_vs_Ridge_PM10_complete}
\small

\begin{tabular}{lllcccccc}
\toprule

\textbf{Station} & \textbf{Scenario} & \textbf{Best ML} &
\multicolumn{3}{c}{\textbf{Best ML}} &
\multicolumn{3}{c}{\textbf{Ridge}} \\
\cmidrule{4-9}
 & & \textbf{Model} & \textbf{RMSE} & \textbf{MAE} & \boldmath{$R^2$} & \textbf{RMSE} & \textbf{MAE} & \boldmath{$R^2$} \\

\midrule
\multirow{6}{*}{Marylebone Road} & Scen1-noT & RF & 8.36 & 6.11 & 0.22 & 9.20 & 6.85 & 0.05 \\
 & Scen1 & RF & 8.43 & 6.27 & 0.20 & 9.20 & 6.84 & 0.05 \\
 & Scen2-noT & LGBM & 5.85 & 4.01 & 0.62 & 6.52 & 4.39 & 0.52 \\
 & Scen2 & LGBM & 5.84 & 4.02 & 0.62 & 6.52 & 4.39 & 0.52 \\
 & Scen3-noT & LGBM & 5.57 & 3.89 & 0.65 & 5.83 & 4.15 & 0.62 \\
 & Scen3 & XGB & 5.52 & 3.95 & 0.66 & 5.86 & 4.17 & 0.62 \\
\midrule
\multirow{6}{*}{Camden Kerbside} & Scen1-noT & RF & 11.34 & 7.21 & 0.14 & 11.61 & 7.23 & 0.09 \\
 & Scen1 & XGB & 10.53 & 6.68 & 0.25 & 11.54 & 7.19 & 0.10 \\
 & Scen2-noT & LGBM & 9.72 & 4.60 & 0.36 & 9.70 & 4.53 & 0.37 \\
 & Scen2 & XGB & 9.70 & 4.54 & 0.37 & 9.74 & 4.57 & 0.36 \\
 & Scen3-noT & LGBM & 9.31 & 4.14 & 0.42 & 9.15 & 4.22 & 0.44 \\
 & Scen3 & XGB & 9.23 & 4.11 & 0.43 & 9.21 & 4.28 & 0.43 \\
\midrule
\multirow{6}{*}{Putney High Street} & Scen1-noT & LGBM & 12.09 & 6.56 & 0.09 & 12.31 & 6.86 & 0.06 \\
 & Scen1 & RF & 12.05 & 6.65 & 0.10 & 12.28 & 6.84 & 0.06 \\
 & Scen2-noT & LGBM & 10.35 & 4.59 & 0.34 & 10.24 & 4.80 & 0.35 \\
 & Scen2 & LGBM & 10.33 & 4.61 & 0.34 & 10.24 & 4.78 & 0.35 \\
 & Scen3-noT & LGBM & 10.44 & 4.14 & 0.32 & 10.55 & 4.74 & 0.31 \\
 & Scen3 & RF & 10.51 & 4.14 & 0.31 & 10.55 & 4.74 & 0.31 \\
\bottomrule
\end{tabular}

\end{table}
\unskip

%------------------------
\begin{table}[H]
\centering
\caption{\PMtwofive model performance: RMSE, MAE, and $R^2$ for each station and scenario for the best-performing tree-based model and the Ridge regression model.}
\label{tab:ML_vs_Ridge_PM25_complete}
\small

\begin{tabular}{lllcccccc}
\toprule

\textbf{Station} & \textbf{Scenario} & \textbf{Best ML} &
\multicolumn{3}{c}{\textbf{Best ML}} &
\multicolumn{3}{c}{\textbf{Ridge}} \\
\cmidrule{4-9}
 & & \textbf{Model} & \textbf{RMSE} & \textbf{MAE} & \boldmath{$R^2$} & \textbf{RMSE} & \textbf{MAE} & \boldmath{$R^2$} \\

\midrule
\multirow{6}{*}{Marylebone Road} & Scen1-noT & LGBM & 4.38 & 3.30 & 0.18 & 4.55 & 3.24 & 0.12 \\
 & Scen1 & XGB & 4.54 & 3.44 & 0.12 & 4.61 & 3.28 & 0.09 \\
 & Scen2-noT & ETR & 3.37 & 2.44 & 0.52 & 3.39 & 2.55 & 0.51 \\
 & Scen2 & RF & 3.32 & 2.44 & 0.53 & 3.49 & 2.65 & 0.48 \\
 & Scen3-noT & LGBM & 3.46 & 2.52 & 0.49 & 3.36 & 2.58 & 0.52 \\
 & Scen3 & RF & 3.38 & 2.44 & 0.51 & 3.41 & 2.63 & 0.50 \\
\midrule
\multirow{6}{*}{Camden Kerbside} & Scen1-noT & XGB & 5.49 & 4.19 & $-$0.62 & 4.23 & 2.96 & 0.04 \\
 & Scen1 & XGB & 4.93 & 3.92 & $-$0.31 & 4.22 & 2.95 & 0.04 \\
 & Scen2-noT & ETR & 2.43 & 1.88 & 0.68 & 2.60 & 2.05 & 0.64 \\
 & Scen2 & ETR & 2.41 & 1.89 & 0.69 & 2.64 & 2.09 & 0.63 \\
 & Scen3-noT & ETR & 2.42 & 1.88 & 0.69 & 2.55 & 2.02 & 0.65 \\
 & Scen3 & RF & 2.54 & 1.95 & 0.65 & 2.57 & 2.04 & 0.64 \\
\bottomrule
\end{tabular}

\end{table}

\subsection{Model Validation for Different Target Points with a Cross-Site~Model}

To assess the coverage capacity of the proposed models, a~validation experiment was conducted using two monitoring stations: Westminster--Oxford Street (hereafter referred to as Oxford Street) and Camden--Euston Road (hereafter referred to as Euston), located in proximity to Marylebone Road, where a model had previously been trained. This experiment was not intended to replace a locally trained model; rather, it aimed to provide an indication of the effective coverage range of the proposed framework in~situations where insufficient historical data are available for local training.

For each validation station, the~predictor sets were constructed following the same methodology described in Section~\ref{sec:Methods} for the six scenarios. Owing to the proximity of the validation stations to Marylebone Road, located approximately 0.96 km from Euston and 1.89 km from Oxford Street, the~neighbouring monitoring stations used in Scen~2 and Scen~3 for both validation locations were the same as those used to train the Marylebone Road models.
The data from the two validation stations were collected between 9 July 2025 and 11 December 2025. This period is about two weeks shorter than the range used for the main analysis, and this section uses this reduced period consistently. 

\begin{figure}[H]
  \centering
  \includegraphics[width=0.99\linewidth]{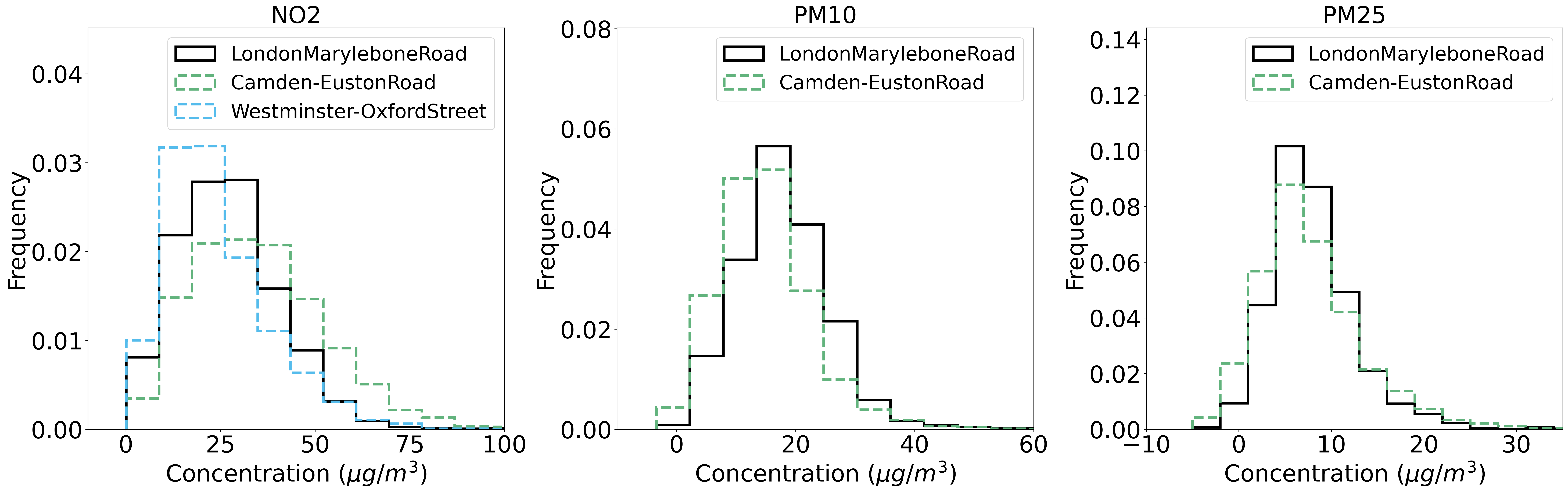}
  \caption{Distribution of pollutant concentrations at the validation and training~stations.}
  \label{fig:distributions_validation}
\end{figure}
\vspace{-6pt}

\begin{figure}[H]
  \centering
  \includegraphics[width=.85\linewidth]{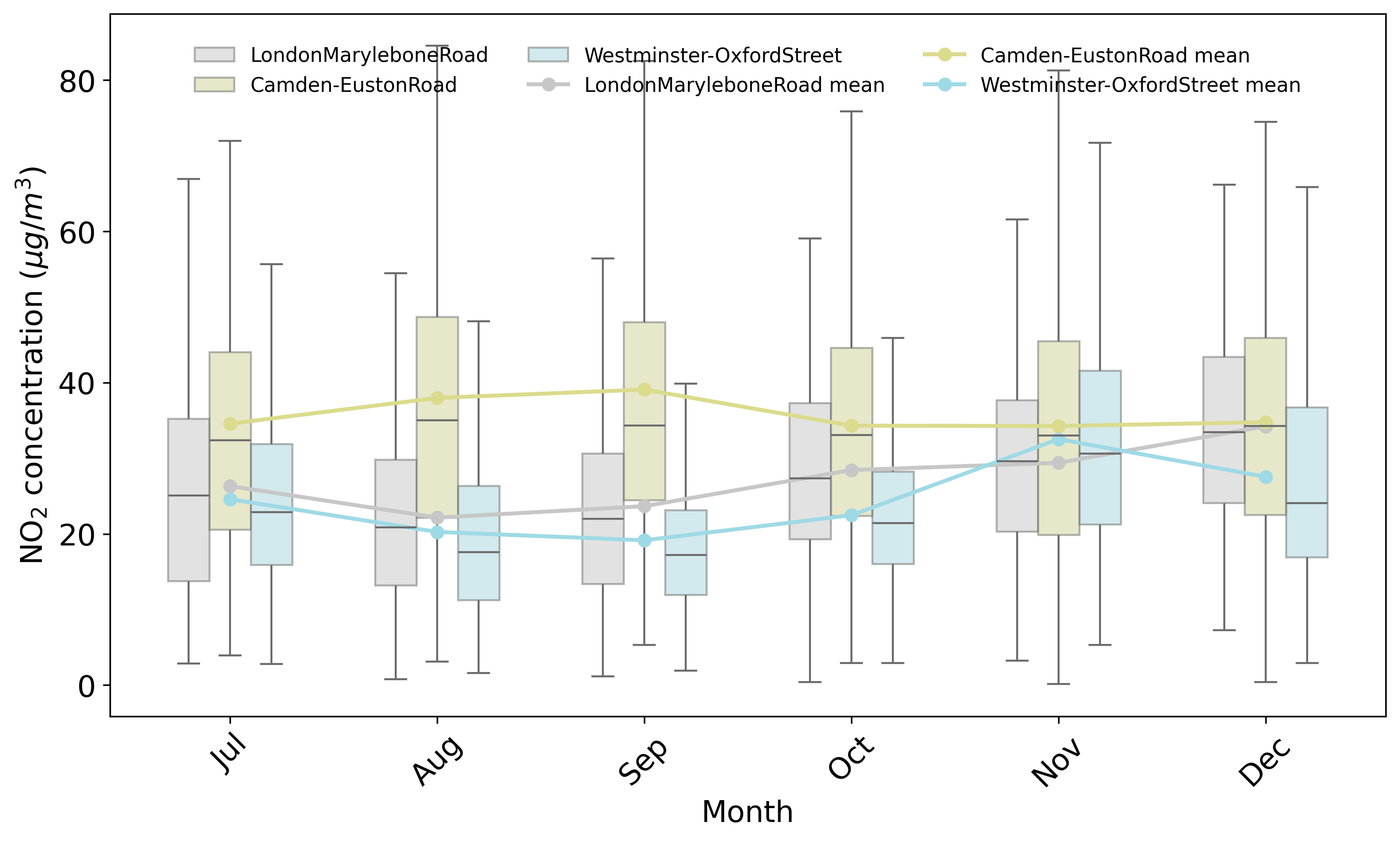}
  \caption{Monthly distribution of \NOtwo concentrations across monitoring stations. Boxplots represent the concentration variability for each month, while points and connecting lines indicate the monthly mean concentrations at each~station.}
  \label{fig:Boxplot_validation}
\end{figure}

\begin{figure}[H]
  \centering
  \includegraphics[width=.85\linewidth]{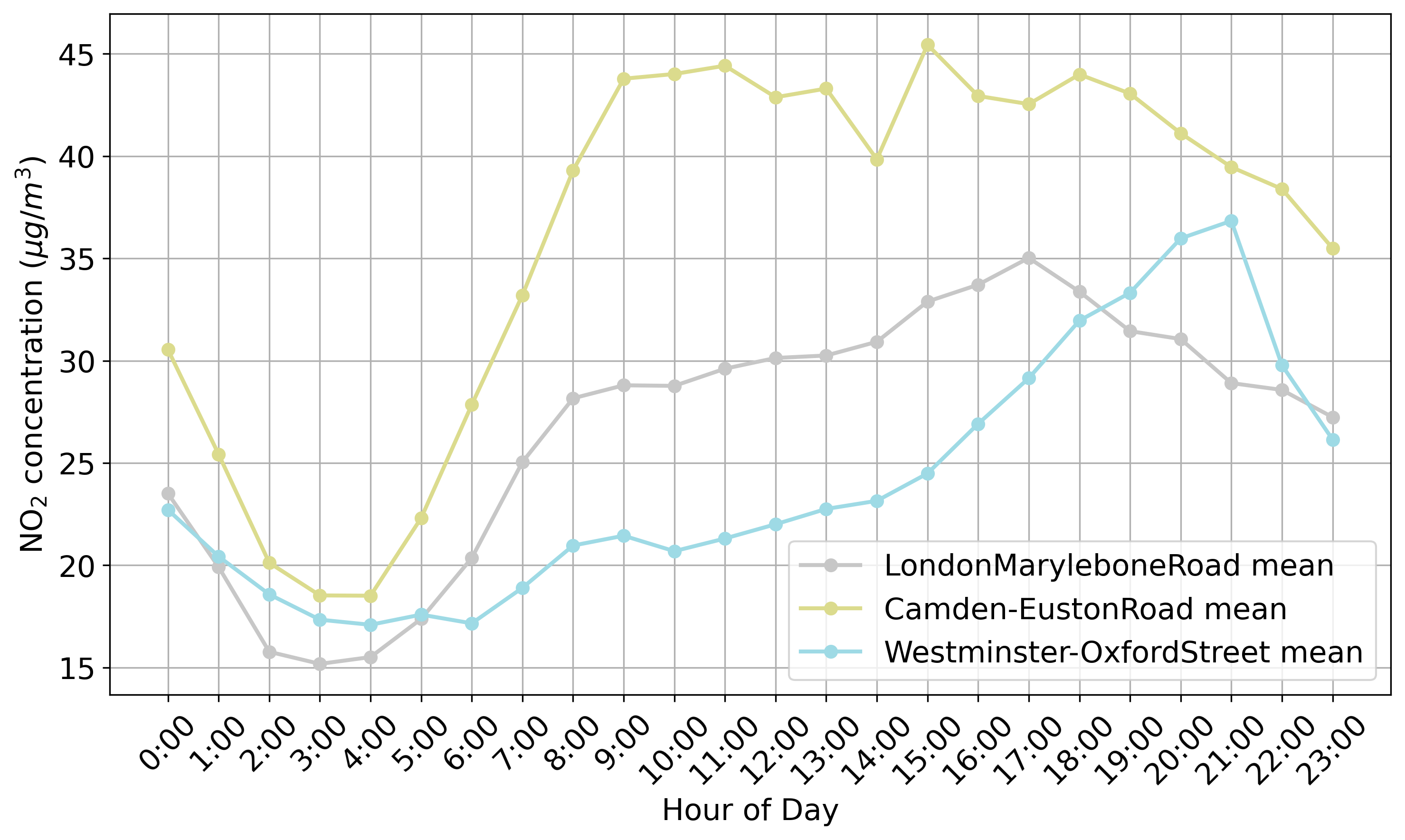}
  \caption{Mean hourly \NOtwo concentrations throughout the study period at Marylebone Road and the validation~stations.}
  \label{fig:hourly_mean_validation}
\end{figure}

Figure~\ref{fig:distributions_validation} presents the distribution of pollutant concentrations at both the validation and training stations, illustrating the similarity of the pollutant patterns for this analysis. Figure~\ref{fig:Boxplot_validation} displays the monthly distribution of \NOtwo concentrations across Marylebone Road and the validation stations, while Figure~\ref{fig:hourly_mean_validation} presents the mean hourly \NOtwo concentrations during the study~period.

To quantify the performance degradation associated with applying a model trained at one monitoring location to a different site, we produce five \NOtwo estimates in the following~configurations:
\begin{itemize}
    \item \textit{Local models}: the machine-learning models were trained independently for each validation station following the same methodology described in Section~\ref{sec:Machine-learning-workflow}.
    \item \textit{Cross-site models}: the models trained using data from Marylebone Road were applied to the data from each validation station.
    \item \textit{Interpolation methods}: for comparison, \NOtwo estimates were also obtained for the same timestamps using the two interpolation methods described in Section~\ref{sec:interpolation}.
\end{itemize}

The observed \NOtwo concentrations were compared
with the predictions generated by the local models, the~cross-site model, and~the interpolation methods using the same independent test set from the fifth chronological split to ensure a fair, paired~comparison.

The local model for Oxford Street achieves RMSE improvements of 12.37\% when comparing Scen~2 with Scen~1, 12.31\% when comparing Scen~3 with Scen~1.
For the Euston local model, the~corresponding improvements are 2.77\%, and~15.81\% respectively. Notably, the~Euston model under Scen~2 exhibits very little improvement from incorporating \NOtwo measurements from the nearest background monitoring station compared with using only the predictors available in Scen~1. However, increasing the predictor set to the complexity of Scen~3 yields an additional improvement of over 10\%.

Table~\ref{tab:cross_site_interpolation_results} presents the RMSE obtained for each station, scenario and method considered in the validation process, together with the performance degradation associated with using the cross-site model instead of the local model, Ridge regression, IDW and OK. 
The performance degradation is quantified using Equation~(\ref{eq:degradation}),  with the local model as the reference scenario (A) and the cross-site model as the alternative scenario (B). Then, positive values indicate a percentage
increase in RMSE when using the cross-site model relative to the corresponding comparison method, representing a loss in predictive performance, whereas negative values indicate a percentage decrease in RMSE, representing a performance advantage of the cross-site model over the comparison~method.

Oxford Street has a lower RMSE achieved by the local models (9.76~\umg on average across the six scenarios), compared with 13.62~\umg at Euston. However, the~cross-site models achieve higher average RMSE values at Oxford Street (14.22~\umg) than at Euston (12.12~\umg). Indeed, when evaluated on a scenario-by-scenario basis, the~cross-site models consistently produce higher RMSE values at Oxford Street than at~Euston.

\begin{table}[H]
\centering
\caption{Performance comparison of the cross-site model with the local model (L), Ridge regression (R), inverse distance weighting (IDW), and~Ordinary Kriging (OK) for NO$_2$ estimation.}
\label{tab:cross_site_interpolation_results}

\resizebox{\textwidth}{!}{%
\begin{tabular}{llccccccccc}
\toprule
\textbf{Station} & \textbf{Scenario} & \textbf{RMSE\boldmath{$_{Cs}$}} & \textbf{RMSE\boldmath{$_L$}} &  \textbf{RMSE\boldmath{$_{R}$}} & \textbf{RMSE\boldmath{$_{IDW}$}} & \textbf{RMSE\boldmath{$_{OK}$}} & \textbf{\boldmath{$D_g$} Cs vs L (\%)} & \textbf{\boldmath{$D_g$} Cs vs R (\%)} & \textbf{\boldmath{$D_g$} Cs vs IDW (\%) } & \textbf{\boldmath{$D_g$} Cs vs OK (\%)} \\
\midrule
Euston & S1 & 12.59 & 14.33 & 14.39 & 18.67 & 16.09 & $-$12.14 & $-$12.51 & $-$32.57 & $-$21.74 \\
Euston & S1-noT & 13.58 & 12.84 & 13.09 & 18.67 & 16.09 & 5.76 & 3.74 & $-$27.26 & $-$15.58 \\
Euston & S2 & 12.24 & 13.20 & 13.85 & 18.67 & 16.09 & $-$7.27 & $-$11.62 & $-$34.44 & $-$23.91 \\
Euston & S2-noT & 12.67 & 13.11 & 12.10 & 18.67 & 16.09 & $-$3.36 & 4.71 & $-$32.14 & $-$21.24 \\
Euston & S3 & 10.60 & 15.28 & 12.62 & 18.67 & 16.09 & $-$30.63 & $-$16.01 & $-$43.22 & $-$34.11 \\
Euston & S3-noT & 11.03 & 12.98 & 10.03 & 18.67 & 16.09 & $-$15.02 & 9.97 & $-$40.92 & $-$31.43 \\
Oxford St & S1 & 15.19 & 10.74 & 12.48 & 13.01 & 13.12 & 41.43 & 21.71 & 16.76 & 15.76 \\
Oxford St & S1-noT & 15.62 & 11.42 & 13.38 & 13.01 & 13.12 & 36.78 & 16.74 & 20.06 & 19.04 \\
Oxford St & S2 & 13.31 & 9.01 & 9.36 & 13.01 & 13.12 & 47.72 & 42.20 & 2.31 & 1.43 \\
Oxford St & S2-noT & 14.02 & 9.88 & 10.41 & 13.01 & 13.12 & 41.90 & 34.68 & 7.76 & 6.84 \\
Oxford St & S3 & 13.32 & 8.24 & 9.31 & 13.01 & 13.12 & 61.65 & 43.07 & 2.38 & 1.51 \\
Oxford St & S3-noT & 13.87 & 9.30 & 9.55 & 13.01 & 13.12 & 49.14 & 45.24 & 6.61 & 5.70 \\
\bottomrule
\end{tabular}
}

\noindent{\footnotesize{Note: $D_g$ is the degradation percentage calculated using Equation~(\ref{eq:degradation}), with the local model as the reference scenario (A) and the cross-site model as the alternative scenario (B). Hence, positive $D_g$ values indicate a percentage increase in RMSE for the cross-site model relative to the corresponding comparison method, whereas negative values indicate a percentage decrease relative to the comparison method, representing a performance advantage of the cross-site model.}}

\end{table}

On the other hand, at~Euston Road, the~cross-site model frequently achieve lower RMSE values than the locally trained models, Ridge regression, and~both interpolation methods, with~reductions ranging from 3.36\% to 43.22\%. The~largest improvement relative to the local model is observed for Scen~3, where the RMSE decreased from 15.28 for the local model to 10.60 for the cross-site model. In~contrast, at~Oxford Street, the~cross-site model generally shows a performance degradation relative to the comparison methods across all scenarios. Specifically, when compared with the locally trained model, increases in RMSE range from 36.78\% to 61.65\%.

The contrasting results between Euston Road and Oxford Street suggest that cross-site transferability is strongly dependent on the target location. Notably, the~model trained at Marylebone Road outperforms the locally trained model at Euston Road in five of the six scenarios, showing that training a model directly at the target site does not necessarily guarantee superior performance on a future independent period. This may indicate that, for~Euston Road, the~relationships learned from the Marylebone Road data generalise better to the independent test period than those learned from the target-site training data, whereas the degradation observed at Oxford Street suggests that such transferability cannot be assumed across~locations. 

Although these findings cannot be generalised given the limited number of target sites, several factors may contribute to the observed differences in model transferability, including road characteristics (Marylebone Road, Oxford Street, and~Euston Road have six, two, and~five traffic lanes, respectively), vehicle fleet composition, traffic patterns, meteorological conditions, nearby pollution sources, and~local urban characteristics. Although~Oxford Street is approximately half as far from Marylebone Road as Euston Road, these results suggest that geographical proximity alone may not determine model transferability and that similarities between the source and target locations may also play an important role.

\subsection{Statistical~Analysis}

Tables ~\ref{tab:bootstrap_scen1}--\ref{tab:bootstrap_scen3} summarise the paired 24-h block-bootstrap comparisons between scenarios with and without traffic-related variables, evaluated on the independent chronological test set. For~each comparison, the~percentage improvement ($\Delta$RMSE), as described in \mbox{Equation~(\ref{eq:percentage_improvement})}, its 95\% confidence interval, and~the Holm-adjusted \emph{p}-value are reported. Positive values of $\Delta$RMSE indicate that the inclusion of traffic-related variables reduced the RMSE. A~value of Yes in the Significant improvement column indicates that the inclusion of traffic-related variables resulted in both a positive RMSE improvement and a statistically significant difference after Holm~correction.

\begin{table}[H]
\centering
\caption{Paired 24-h block-bootstrap comparison between Scen~1 and Scen~1-noT.}
\label{tab:bootstrap_scen1}
\resizebox{\textwidth}{!}{%
\begin{tabular}{lllclcc}
\toprule
\textbf{Pollutant} & \textbf{Station} & \textbf{Best Model} & \boldmath{$\Delta\text{RMSE}$} \textbf{(\%)} & \textbf{95\% CI} & \textbf{Holm-Adjusted \boldmath{$p$}-Value} & \textbf{Significant Improvement} \\
\midrule
NO$_2$ & Marylebone Road & XGB & 10.38 & [0.56, 16.70] & 0.058 & No \\
NO$_2$ & Camden Kerbside & XGB & 1.46 & [$-$0.37, 3.08] & 0.214 & No \\
NO$_2$ & Putney High Street & XGB & 1.21 & [$-$2.07, 4.57] & 0.471 & No \\
NO$_2$ & Oxford Street & XGB & 5.95 & [0.89, 10.18] & 0.030 & Yes \\
NO$_2$ & Euston Road & ETR & $-$11.61 & [$-$21.48, $-$1.96] & 0.072 & No \\
O$_3$ & Marylebone Road & LGBM & $-$5.79 & [$-$10.71, $-$0.85] & 0.045 & No \\
PM$_{10}$ & Marylebone Road & RF & $-$0.90 & [$-$3.35, 1.09] & 1.000 & No \\
PM$_{10}$ & Camden Kerbside & XGB & 7.14 & [4.39, 11.72] & 0.001 & Yes \\
PM$_{10}$ & Putney High Street & RF & 0.33 & [$-$1.88, 1.50] & 1.000 & No \\
PM$_{2.5}$ & Marylebone Road & XGB & $-$3.64 & [$-$8.61, 1.37] & 0.615 & No \\
PM$_{2.5}$ & Camden Kerbside & XGB & 10.19 & [2.66, 16.92] & 0.032 & Yes \\
\bottomrule
\end{tabular}
}

\noindent\parbox{\linewidth}{\footnotesize{Note: 
 $\Delta\text{RMSE}(\%)$ is defined in Equation~(\ref{eq:percentage_improvement}), where A represents the scenario without traffic-related variables, and~B represents the corresponding scenario including traffic-related variables. Positive values indicate that including traffic variables reduced the RMSE. A~statistically significant improvement is reported as Yes only when $\Delta_{\text{RMSE}}>0$ and the Holm-adjusted $p$-value is below 0.05.}}
\end{table}
\unskip

\begin{table}[H]
\centering
\caption{Paired 24-h block-bootstrap comparison between Scen~2 and Scen~2-noT.}
\label{tab:bootstrap_scen2}
\resizebox{\textwidth}{!}{%
\begin{tabular}{lllclcc}
\toprule
\textbf{Pollutant} & \textbf{Station} & \textbf{Best Model} & \boldmath{$\Delta\text{RMSE}$} \textbf{(\%)} & \textbf{95\% CI} & \textbf{Holm-Adjusted \boldmath{$p$}-Value} & \textbf{Significant Improvement} \\
\midrule
NO$_2$ & Marylebone Road & XGB & 9.28 & [2.62, 14.21] & 0.020 & Yes \\
NO$_2$ & Camden Kerbside & XGB & 0.91 & [$-$2.14, 3.91] & 0.555 & No \\
NO$_2$ & Putney High Street & LGBM & 4.53 & [1.71, 7.78] & 0.007 & Yes \\
NO$_2$ & Oxford Street & LGBM & 8.78 & [3.93, 12.98] & 0.002 & Yes \\
NO$_2$ & Euston Road & XGB & $-$0.71 & [$-$9.22, 7.36] & 0.873 & No \\
O$_3$ & Marylebone Road & LGBM & $-$2.12 & [$-$5.48, 1.62] & 0.233 & No \\
PM$_{10}$ & Marylebone Road & LGBM & 0.26 & [$-$1.18, 2.06] & 1.000 & No \\
PM$_{10}$ & Camden Kerbside & XGB & 0.14 & [$-$3.02, 1.45] & 0.875 & No \\
PM$_{10}$ & Putney High Street & LGBM & 0.13 & [$-$1.05, 0.70] & 1.000 & No \\
PM$_{2.5}$ & Marylebone Road & RF & 1.32 & [$-$4.20, 6.70] & 1.000 & No \\
PM$_{2.5}$ & Camden Kerbside & ETR & 0.63 & [$-$1.16, 2.61] & 0.500 & No \\
\bottomrule
\end{tabular}
}
\end{table}
\unskip

\begin{table}[H]
\centering
\caption{Paired 24-h block-bootstrap comparison between Scen~3 and Scen~3-noT.}
\label{tab:bootstrap_scen3}
\resizebox{\textwidth}{!}{%
\begin{tabular}{lllclcc}
\toprule
\textbf{Pollutant} & \textbf{Station} & \textbf{Best Model} & \boldmath{$\Delta\text{RMSE}$} \textbf{(\%)} & \textbf{95\% CI} & \textbf{Holm-Adjusted \boldmath{$p$}-Value} & \textbf{Significant Improvement} \\
\midrule
NO$_2$ & Marylebone Road & LGBM & 5.44 & [$-$1.50, 10.59] & 0.093 & No \\
NO$_2$ & Camden Kerbside & LGBM & 4.63 & [1.09, 8.11] & 0.043 & Yes \\
NO$_2$ & Putney High Street & XGB & 6.86 & [4.08, 9.20] & <0.001 & Yes \\
NO$_2$ & Oxford Street & XGB & 11.42 & [5.30, 17.34] & 0.002 & Yes \\
NO$_2$ & Euston Road & LGBM & $-$17.67 & [$-$31.94, $-$2.92] & 0.072 & No \\
O$_3$ & Marylebone Road & XGB & 4.21 & [1.11, 6.64] & 0.027 & Yes \\
PM$_{10}$ & Marylebone Road & XGB & 0.91 & [$-$2.15, 3.83] & 1.000 & No \\
PM$_{10}$ & Camden Kerbside & XGB & 0.88 & [$-$0.81, 1.37] & 0.438 & No \\
PM$_{10}$ & Putney High Street & RF & $-$0.68 & [$-$2.45, 0.95] & 0.798 & No \\
PM$_{2.5}$ & Marylebone Road & RF & 2.30 & [$-$2.31, 7.09] & 1.000 & No \\
PM$_{2.5}$ & Camden Kerbside & RF & $-$5.14 & [$-$8.95, $-$2.07] & 0.032 & No \\
\bottomrule
\end{tabular}
}
\end{table}

In Scen 1, statistically significant improvements after including traffic-related variables are observed for \NOtwo at Oxford Street, \PMten at Camden Kerbside, and~\PMtwofive at Camden Kerbside, with~RMSE reductions ranging from approximately 5.95\% to 10.19\%. 

Scen 2 provides the most consistent benefit from incorporating traffic-related variables. Statistically significant RMSE reductions are obtained for \NOtwo at Marylebone Road, Putney High Street, and~Oxford Street, ranging from approximately 4.53\% to 9.28\%. In~contrast, no statistically significant improvements are observed for \Othree, \PMten, or~\PMtwofive.

In Scen 3, significant improvements are again observed for \NOtwo, particularly at Camden Kerbside, Putney High Street, and~Oxford Street, together with \Othree~at Marylebone Road, ranging from 4.63\% to 11.42\%. Conversely, the~inclusion of traffic-related variables significantly degrades performance for \PMtwofive at Camden Kerbside, while \NOtwo~at Euston Road also exhibits a substantial RMSE~increase.

Overall, the~statistical analysis indicates that the predictive benefit of traffic-related variables is heterogeneous across pollutants, stations, and~feature scenarios, rather than providing a uniform improvement across all cases. Improvements are concentrated mainly in \NOtwo~predictions, particularly under Scen~2 and Scen~3, whereas \PMten shows only isolated gains and \PMtwofive~exhibits mixed behaviour, including one statistically significant deterioration. \Othree~shows a statistically significant improvement only at Marylebone Road under Scen~3; however, as~this is the only traffic-type station where \Othree~measurements are available, this result cannot be generalised to other~sites.

The statistically significant improvements identified in this study are associated with RMSE reductions of approximately 4.5–11.4\%, indicating non-negligible improvements in predictive performance. Nevertheless, these benefits are not consistent across pollutants and sites, and~statistically significant performance degradations are also observed in some cases. These results highlight the need for further research to identify the site characteristics and local conditions under which traffic-related data provide a meaningful predictive~benefit.

\section{Discussion}
\label{sec:Discussion}
\unskip
\subsection{Understanding the Role of Traffic and Spatial~Context}

The performance of the models was evaluated by comparing the three base predictor sets (scenarios) with their corresponding configurations after removing the traffic-related variables to assess whether their inclusion is worthwhile in the design of air-quality monitoring systems. These findings extend those reported by ref.~\cite{Sulaimon2022}, in~which only a configuration equivalent to Scen~1 was analysed, and~by ref.~\cite{Samad2023}, which compared %compares 
scenarios with an increasing number of predictors, without~isolating the specific contribution of traffic-related~variables. 

As expected, the~contribution of traffic-related variables is found to be pollutant- and site-dependent, with~greater improvements observed at some monitoring locations than at~others.

It is reported in ref.~\cite{DEVEER2025} that \PMtwofive was heavily influenced by the traffic information represented as traffic emissions and speed, while \NOtwo showed better results when traffic was represented as vehicle counts. In~contrast, ref.~\cite{Samad2023} used vehicle counts as a traffic descriptor, and there were no significant improvements reported in the RMSE of \PMtwofive and \PMten estimates when traffic information was incorporated alongside meteorological variables and data from one air-quality monitoring station (the equivalent of Scen~2 in this study). Instead, the~greatest contribution of traffic data was observed for \NOtwo prediction.
In the present study, \NOtwo~exhibited the greatest improvement from incorporating traffic variables based on travel time and speed, showing the highest average RMSE improvement across the three scenarios. In~contrast, \PMten, \PMtwofive, and~\Othree~showed heterogeneous responses to the inclusion of traffic-related variables, with~RMSE improvements in some scenarios and performance deterioration in~others.

One possible explanation for why traffic-related variables did not improve \Othree~prediction in the simpler scenarios to the same extent as for \NOtwo could be that \Othree~is a secondary pollutant, and~traffic variables therefore do not directly represent its formation processes. Interestingly, the~contribution of traffic-related variables became beneficial under Scen~3, where measurements of pollutants involved in \Othree chemistry, including \NOtwo, NO, and~NO$_\text{x}$ as \NOtwo, were also available as predictors. This may indicate that traffic information becomes more informative for \Othree~prediction when considered together with variables related to its photochemical formation and titration processes. Nevertheless, \Othree~was evaluated at only one station, which again limits the interpretation of these results.

Similar to the conclusions in ref.~\cite{Samad2023}, in~this study, when comparing predictive performance across scenarios, the~inclusion of pollutant measurements from nearby stations also led to lower RMSE values than those obtained using traffic, temporal, and~meteorological variables alone. 

The SHAP analysis conducted in the present study showed that, at~least for \NOtwo and \Othree, traffic-related variables played an important role in prediction, in~some cases even greater than that of other pollutants measured at auxiliary stations. Therefore, as~future work, it would be worthwhile to investigate whether some pollutants measured at auxiliary stations could be excluded from Scen~3 in order to reduce the data requirement associated with predicting \NOtwo or \Othree, once measurements of the same pollutant from nearby stations, together with local meteorological and traffic information, are already available.
In contrast, for~\PMtwofive, traffic-related variables did not have an impact on improving the predictions; in some cases, they even produced worse results, suggesting that, rather than contributing useful information, they may introduce noise into the modelling process.

\subsection{Data~Requirement--Performance Trade-Offs in Monitoring~Design}

In ref.~\cite{Samad2023}, a~configuration comparable to Scen~1 without traffic was found to perform worse than Scen~3 with traffic. Similarly, in~the present study, Scen~3 outperformed Scen~1 across all analysed locations and pollutants. However, the~comparison between Scen~3 and Scen~2 showed performance deterioration in some cases and only small improvements in others, highlighting that differences in model performance between scenarios should also be evaluated in terms of data requirement--performance trade-offs. This involves assessing not only the contribution of traffic-related variables but also whether the additional information required from multiple air-quality monitoring stations provides sufficient predictive improvement to justify the increased data~requirements.

For \PMten estimation at the three analysed sites, the~RMSE improvement obtained when moving from measurements from one neighbouring monitoring station (Scen~2) to measurements from four neighbouring stations (Scen~3) was below 5.37\%, including one case of performance deterioration. For~\PMtwofive, both analysed cases showed an increase in RMSE when incorporating data from the additional neighbouring stations. These results indicate that increasing the amount of monitoring-station information available to the model does not necessarily translate into better predictive~performance.

The results from the validation experiment suggest that cross-site models may represent a practical alternative for extending air-quality estimation to some locations. However, given the limited number of cases analysed, it is not possible to identify which characteristics of the target and training locations may be relevant to successful model transferability. Although some degradation in performance relative to locally trained models may be expected, cross-site models can, in~some cases, provide estimates with a reasonable degree of accuracy. Consequently, the~proposed approach may contribute to improving access to air-quality information in areas where monitoring stations are unavailable or sparsely distributed, but~where traffic and meteorological data can be obtained remotely through publicly available or commercial data~services.

These results are not intended to suggest that the proposed approach can replace fixed-site air-quality monitoring networks, which, as~discussed previously, remain the most reliable source of accurate pollutant measurements. Rather, the~proposed framework aims to complement existing monitoring infrastructure by exploiting the available information from neighbouring monitoring stations together with traffic, temporal, and~meteorological variables for air pollutant~estimation.

\subsection{Scalability and Spatial Deployment~Considerations}

The proposed framework relies on already existing infrastructure and does not require the installation of a new physical monitoring station. Nevertheless, acquiring data through commercial API services presents certain limitations. In~the case of commercial providers, the number of free API requests per day is typically restricted to a relatively small amount, whereas subscriptions to paid plans are limited by institutional budgets. Under~paid schemes, pricing depends on both the number of queried locations and the sampling frequency requested, with~traffic-related data being particularly expensive and not obviously useful in terms of predictive value for air pollution~estimation.

Hence, in~operational settings, especially for practical large-scale deployment, the~framework would require careful query-point selection for traffic and meteorological data acquisition. Therefore, future work should investigate optimal spatial querying strategies capable of identifying representative urban locations while minimising operational costs. Potential approaches may include Voronoi-based spatial partitioning, minimum-distance criteria between sampled locations, or~spatial coverage optimisation methods considering road-network segmentation and urban~morphology.

\section{Conclusions} 
\label{sec:Conclusions}
This work contributes to a growing body of evidence supporting the use of remote, commercially accessible traffic data as a meaningful input for ML-based urban air quality modelling. By~systematically evaluating the contribution of traffic information across six predictor scenarios and five target locations in London, we reach conclusions that are relevant both to the predictive modelling literature and to practitioners involved in air quality monitoring network~design. 

The central finding of this study is that traffic-related information, such as speed and travel time, together with meteorological data, can provide additional predictive value for the estimation of \NOtwo, \Othree, and~\PMten. The~SHAP analysis revealed that traffic-related variables can contribute at a level comparable to, and~in some cases greater than, measurements of the target pollutant obtained from neighbouring monitoring stations. This behaviour was particularly evident under Scen~2 at Marylebone Road and Putney High Street, where the contribution of traffic-related variables was similar to that of the nearest auxiliary-station \NOtwo measurements. Furthermore, under~Scen~3 for \NOtwo prediction, traffic-related variables ranked above auxiliary pollutant measurements (aside from \NOtwo) in terms of feature importance across the three stations considered in the main analysis. These findings are further supported by the Euston validation experiment, where incorporating \NOtwo measurements from the nearest monitoring station (Scen~2) provided only a marginal improvement over the traffic-, temporal-, and~meteorology-based configuration (Scen~1), suggesting that traffic-state information may partially compensate for the absence of nearby air-quality observations under certain urban~conditions.

The results from the validation experiment suggest that cross-site models can, in~some cases, provide a practical alternative for locations where historical air-pollution measurements are unavailable but real-time traffic, temporal, and~meteorological data can be obtained. 
Although these results cannot be generalised to other sites, performance degradation relative to locally trained models was observed in this study, ranging from 5.76\% to 61.65\% across the evaluated scenarios. Conversely, some scenarios showed improved performance, with~improvements ranging from 3.36\% to 30.63\%. The~resulting prediction errors remained within a range that may still be useful for some air-quality assessment applications, for~example, at~locations lacking monitoring~infrastructure.

Nevertheless, given the limited number of sites analysed, it is not possible to determine which characteristics of the source and target locations favour successful model transferability. The~observed results suggest that geographical proximity alone may not be sufficient and that factors such as urban morphology, road characteristics, nearby emission sources, and~local environmental conditions may also be relevant. Consequently, locally trained models remain preferable whenever sufficient historical observations are available, whereas cross-site models may offer a valuable alternative in data-scarce environments.

A second contribution is methodological: the combination of a scenario-based ablation design with four tree-based ensemble models with classical interpolation and regression baselines enables a controlled decomposition of information gain that has been absent from prior comparative studies. Unlike works that evaluate traffic as a binary predictor (present or absent) this framework quantifies how the relative importance of traffic shifts as the information environment changes, offering a transferable evaluation protocol for future studies. By~using this framework, we found that traffic data can provide valuable complementary information, particularly in environments where traffic is a predominant source of pollution. 
Nevertheless, its contribution was highly context-dependent, and~pollutant measurements from traffic and background stations, together with meteorological variables, generally exhibited greater predictive value and feature~importance.

From a practical perspective, the~proposed framework relies on publicly accessible or commercially available datasets that offer free request allowances, although~subject to daily or monthly request limits. This allows the framework to be reproduced and deployed without the need for additional sensing~infrastructure. 

The reliance on these data sources (e.g., UK-AIR, OpenWeather, and~TomTom) makes the proposed pipeline potentially reproducible in cities where equivalent air-quality data and commercial traffic and meteorological data services are available, reducing the barriers to deploying virtual air-quality monitoring systems in low- and middle-income \mbox{urban~contexts.}

It is important to note that the London case study was conducted using monitoring stations located within the Ultra Low Emission Zone (ULEZ). Consequently, traffic composition, fleet characteristics, and~pollutant concentrations may differ from other cities operating under different traffic management and emission-regulation frameworks.

Three directions for future work are identified in order to improve the generalisability of the study: (i) validation of the framework in other cities with different atmospheric dynamics, traffic compositions, and~climate conditions; (ii) integration of spatio-temporal deep learning architectures that can model dependencies across the monitoring network dynamically; and (iii) exploration of emission inventory features at multi-source sites to extend the applicability of the approach beyond traffic-dominated~corridors.

Future work should also extend the cross-site transferability study.
The~cross-site transferability analysis was limited by data availability, which meant that transferability could only be assessed at two additional stations and only for \NOtwo.
Given this small sample, future efforts should evaluate transferability across a larger number of sites and for \mbox{multiple~pollutants.}

The study period was constrained by time and resource limitations associated with the real-time collection of traffic data through the TomTom API, resulting in a dataset covering June to December. Given the relationship between \Othree~and temperature and humidity, its seasonal cycle~\cite{ukozone} could not be fully captured within the study period, limiting the generalisability of its predictions across the year. Therefore, in~future research it would be beneficial to extend the study period to the whole year so that we can validate model performance over the full yearly cycle.

The conclusions of this study, beyond~the specific variables analysed, suggest that incorporating relevant information directly related to pollution sources can improve pollutant prediction, enabling the approximation of their influence without the need for additional sensors or extensive monitoring networks. Hence, consideration could also be given to incorporating information related to other types of pollution sources, as~well as additional data formats such as satellite imagery of atmospheric conditions, traffic flow video data, or~quantitative descriptors of the surrounding environment, including the presence of green areas, nearby industrial facilities, and~other urban~characteristics.

\subsection{Data Availability}
\label{subsec:DataAvailability}
Restrictions apply to the availability of these data. 
Traffic and meteorological data providers do not permit the large-scale redistribution of data downloaded directly from their APIs. Raw traffic data were acquired in real time through the TomTom Flow Segment Data API~\cite{TomTom_Flow_Segment_Data} under its free-access scheme. Meteorological data can be reproduced by querying the OpenWeather One Call API 3.0~\cite{open_weather} under the user’s own credentials and licensing conditions, although~free access is subject to a limited number of daily requests. Historical air pollution data are publicly available from the official UK government repository in~\cite{defra_ukair_data}.
The code is available on Zenodo at DOI: \href{https://zenodo.org/records/21863941}{https://zenodo.org/records/21863941}.

\subsection{Acknowledgements}
\label{subsec:Acknowledgements}
We thank our institutions, the~Instituto Politécnico Nacional (IPN) and Queen Mary University of London (QMUL), for~enabling this collaboration through the provision of IT facilities and access to invaluable expertise, particularly from Daohai Li and Alex Owen at QMUL. We also gratefully acknowledge the sponsors who supported this collaboration, such as Fundación Politécnico and~COFAA-IPN.

\subsection{Author contributions}
Conceptualisation: M.B., V.L.-S., J.R.
Data Curation: V.L.-S.
Formal Analysis: M.B., V.L.-S., J.R.
Funding Acquisition: A.A., M.B., V.L.-S., M.S.-P.
Investigation: M.B., V.L.-S., J.R.
Methodology: M.B., V.L.-S., J.R.
Project Administration: M.B.
Resources: A.A., M.B., M.S.-P.
Software: V.L.-S.
Supervision: A.A., M.B., M.S.-P.
Validation: V.L.-S.
Visualisation: V.L.-S.
Writing---Original Draft Preparation: M.B., V.L.-S.
Writing---Review \& Editing: A.A., M.B., V.L.-S., J.R., M.S.-P. All authors have read and agreed to the published version of the manuscript

\subsection{Funding} 
A.A. received funding from the SIP-IPN under grant agreement No. SIP20250146, SIP-2284 and ProRed 2026-0060. 
M.B. received funding from the Science and Technology Facilities Council (STFC) Impact Acceleration Account.
V.L.-S. receives financial support from the Secretaría de Ciencia, Humanidades, Tecnología e Innovación (SECIHTI) under grant 960525; the Fundación Politécnico, IPN and QMUL through the dual doctoral programme between the IPN and QMUL; and the Comisión de Operación y Fomento de Actividades Académicas (COFAA-IPN). J.R. receives funding from QMUL via a Principal's PhD Studentship.
M.S.-P. has received funding from the SIP-IPN under grants IND-2026-0524, SECTEI/084/2023 and CM-SECTEI/040/2025. M.S.-P. and V.L.-S. received funding from the SIP-IPN under grant~2024-A175.

\subsection{Competing interests}
The authors declare no conflicts of~interest.

\vspace{6pt} 

\bibliographystyle{unsrtnat} % or unsrtnat for order of appearance
\bibliography{bibliography}

\appendix

\section{Supplementary Material}
\label{App}
\renewcommand{\thetable}{S\arabic{table}}
\setcounter{table}{0}

The figures part of the supplementary material are available as ancillary files accompanying this submission.
The supplementary material provides information for all monitoring stations considered in this study, including the validation stations that were analysed separately from the main experimental dataset.

The figures in the ancillary material are listed below.
\begin{description}
    \item[Figure S1] corresponds to the file \texttt{OpenWeather\_validation\_temperature\_wind.png}.
    Comparison of OpenWeather meteorological data with observations from the air quality monitoring stations at Camden Kerbside and Marylebone Road.
    \item[Figure S2] corresponds to the file \texttt{Correlation\_Matrix\_LondonMaryleboneRoad.png}.
    Correlation matrix between temporal, remotely acquired traffic and meteorological variables, and criteria pollutants measured at the London Marylebone Road station.
    \item[Figure S3] corresponds to the file \texttt{Correlation\_Matrix\_Wandsworth-PutneyHighStreet.png}.
    Correlation matrix between temporal, remotely acquired traffic and meteorological variables, and criteria pollutants measured at the Wandsworth – Putney High Street station.
    \item[Figure S4] corresponds to the file \texttt{Correlation\_Matrix\_CamdenKerbside.png}.
    Correlation matrix between temporal, remotely acquired traffic and meteorological variables, and criteria pollutants measured at the Camden Kerbside station.
    \item[Figure S5] corresponds to the file \texttt{Correlation\_Matrix\_Westminster-OxfordStreet.png}.
    Correlation matrix between temporal, remotely acquired traffic and meteorological variables, and criteria pollutants measured at the Westminster – Oxford Street station.
    \item[Figure S6] corresponds to the file \texttt{Correlation\_Matrix\_Camden-EustonRoad.png}.
    Correlation matrix between temporal, remotely acquired traffic and meteorological variables, and criteria pollutants measured at the Camden – Euston Road station.
    \item[Figure S7] corresponds to the file \texttt{time\_series\_observed\_ML\_Ridge\_PM10.png}.
    Observed and predicted $PM_{10}$ concentrations on the final chronological test set, comparing the best-performing tree-based model with Ridge regression.
    \item[Figure S8] corresponds to the file \texttt{time\_series\_observed\_ML\_Ridge\_NO2.png}.
    Observed and predicted $NO_2$ concentrations on the final chronological test set, comparing the best-performing tree-based model with Ridge regression.
    \item[Figure S9] corresponds to the file \texttt{time\_series\_observed\_ML\_Ridge\_O3.png}.
    Observed and predicted $O_3$concentrations on the final chronological test set, comparing the best-performing tree-based model with Ridge regression.
    \item[Figure S10] corresponds to the file \texttt{time\_series\_observed\_ML\_Ridge\_PM25.png}.
    Observed and predicted $PM_{2.5}$ concentrations on the final chronological test set, comparing the best-performing tree-based model with Ridge regression.
\end{description}

\begin{table}[!hbpt]
\centering
\caption{Number of observations available from each data source before and after merging all data sources.}
\label{tab:preprocessing_data_availability}
\resizebox{\textwidth}{!}{%
\begin{tabular}{llrrrrrrrrr}
\toprule
Station & Pollutant & Pollution & Traffic & Pollution + traffic & Weather & Bkg0 & Bkg1 & Trf0 & Trf1 & Final merge \\
\midrule
Euston Road & NO2 & 4152 & 4510 & 3747 & 4151 & 4151 & 4151 & 4151 & 4151 & 3747 \\
Camden Kerbside & NO2 & 4152 & 4706 & 3943 & 4152 & 4151 & 4151 & 4151 & 4151 & 3943 \\
Camden Kerbside & PM10 & 4152 & 4706 & 3943 & 4152 & 4151 & 4151 & 4151 & 4151 & 3943 \\
Camden Kerbside & PM25 & 4152 & 4706 & 3943 & 4152 & 4151 & 4151 & 4151 & 4151 & 3943 \\
Marylebone Road & NO2 & 4152 & 4707 & 3944 & 4152 & 4151 & 4151 & 4151 & 4151 & 3944 \\
Marylebone Road & O3 & 4152 & 4707 & 3944 & 4152 & 4151 & 4151 & 4151 & 4151 & 3944 \\
Marylebone Road & PM10 & 4152 & 4707 & 3944 & 4152 & 4151 & 4151 & 4151 & 4151 & 3944 \\
Marylebone Road & PM25 & 4152 & 4707 & 3944 & 4152 & 4151 & 4151 & 4151 & 4151 & 3944 \\
Putney High Street & NO2 & 4152 & 4510 & 3747 & 4152 & 3747 & 4151 & 4151 & 4151 & 3747 \\
Putney High Street & PM10 & 4152 & 4510 & 3747 & 4152 & 3747 & 4151 & 4151 & 4151 & 3747 \\
Oxford Street & NO2 & 4152 & 4510 & 3747 & 4152 & 4151 & 4151 & 4151 & 4151 & 3747 \\
\bottomrule
\end{tabular}

}
\begin{minipage}{\textwidth}
\footnotesize
\textit{Note:} Pollution and traffic indicate the number of observations available from the target monitoring station and hourly aggregated traffic data, respectively. Bkg0 and Bkg1 denote the two nearest background stations, while Trf0 and Trf1 denote the two nearest traffic stations. Final merge indicates the number of observations remaining after merging all data sources by timestamp.
\end{minipage}
\end{table}

\begin{table}[!hbpt]
\centering
\caption{Number of observations and missing values during data preprocessing and temporal interpolation.}
\label{tab:preprocessing_missing_data}
\resizebox{\textwidth}{!}{%
\begin{tabular}{llrrrrrrrr}
\toprule
Station & Pollutant & After cleaning & Missing target & After target removal & Missing predictors & Rows with missing & Interpolated values & Rows with missing (\%) & Missing values interpolated (\%) \\
\midrule
Euston Road & NO2 & 3747 & 515 & 3232 & 424 & 264 & 236 & 8.17 & 55.66 \\
Camden Kerbside & NO2 & 3943 & 31 & 3912 & 866 & 517 & 443 & 13.22 & 51.15 \\
Camden Kerbside & PM10 & 3943 & 103 & 3840 & 871 & 503 & 449 & 13.10 & 51.55 \\
Camden Kerbside & PM25 & 3943 & 74 & 3869 & 880 & 507 & 464 & 13.10 & 52.73 \\
Marylebone Road & NO2 & 3944 & 53 & 3891 & 612 & 336 & 304 & 8.64 & 49.67 \\
Marylebone Road & O3 & 3944 & 334 & 3610 & 560 & 316 & 293 & 8.75 & 52.32 \\
Marylebone Road & PM10 & 3944 & 243 & 3701 & 570 & 311 & 271 & 8.40 & 47.54 \\
Marylebone Road & PM25 & 3944 & 234 & 3710 & 542 & 295 & 274 & 7.95 & 50.55 \\
Putney High Street & NO2 & 3747 & 10 & 3737 & 1151 & 718 & 456 & 19.21 & 39.62 \\
Putney High Street & PM10 & 3747 & 28 & 3719 & 1139 & 712 & 443 & 19.14 & 38.89 \\
Oxford Street & NO2 & 3747 & 90 & 3657 & 661 & 385 & 323 & 10.53 & 48.87 \\
\bottomrule
\end{tabular}
}
\begin{minipage}{\textwidth}
\footnotesize
\textit{Note:} Missing target indicates observations removed because the target pollutant concentration was unavailable. Missing predictors indicates the total number of missing predictor values before interpolation, whereas Rows with missing indicates the number of observations containing at least one missing predictor. Interpolated values indicates the number of unique missing predictor values recovered by temporal interpolation in at least one of the cross-validation partitions in which they occurred. Rows with missing (\%) indicates the proportion of rows containing at least one missing predictor value, whereas Missing values interpolated (\%) indicates the proportion of missing predictor values recovered through interpolation relative to the total number of missing predictor values.
\end{minipage}
\end{table}

\begin{table}[!htbp]
\centering
\scriptsize
\setlength{\tabcolsep}{4pt}
\renewcommand{\arraystretch}{1.15}

\caption{Hyperparameter search space used for model optimisation. RF: Random Forest; ETR: Extra Trees Regressor; LGBM: LightGBM; XGB: XGBoost.}
\begin{tabular}{p{1.4cm} p{3.2cm} p{6.8cm} p{2.0cm}}
\toprule
Model & Hyperparameter & Search space & Scale / condition \\
\midrule

RF & n\_estimators & [200, 800], step 100 & integer \\
RF & max\_depth & [4, 30] & integer \\
RF & min\_samples\_split & [2, 30] & integer \\
RF & min\_samples\_leaf & [1, 15] & integer \\
RF & max\_features & \{sqrt, log2, 0.3, 0.5, 0.7, 0.9\} & categorical \\
RF & bootstrap & \{True, False\} & categorical \\
RF & ccp\_alpha & [$10^{-6}$, $10^{-2}$] & log-uniform \\
RF & max\_samples & [0.50, 0.95] & if bootstrap = True \\
\midrule

ETR & n\_estimators & [200, 800], step 100 & integer \\
ETR & max\_depth & [4, 35] & integer \\
ETR & max\_leaf\_nodes & [20, 300] & integer \\
ETR & min\_samples\_split & [2, 30] & integer \\
ETR & min\_samples\_leaf & [1, 15] & integer \\
ETR & max\_features & \{sqrt, log2, 0.3, 0.5, 0.7, 0.9\} & categorical \\
ETR & bootstrap & \{True, False\} & categorical \\
ETR & ccp\_alpha & [$10^{-6}$, $10^{-2}$] & log-uniform \\
ETR & max\_samples & [0.50, 0.95] & if bootstrap = True \\
\midrule

LGBM & n\_estimators & [300, 1500], step 100 & integer \\
LGBM & learning\_rate & [$5 \times 10^{-3}$, 0.08] & log-uniform \\
LGBM & num\_leaves & [15, 255] & integer \\
LGBM & max\_depth & [3, 12] & integer \\
LGBM & min\_child\_samples & [10, 120] & integer \\
LGBM & min\_child\_weight & [$10^{-3}$, 10] & log-uniform \\
LGBM & subsample & [0.50, 1.00] & uniform \\
LGBM & subsample\_freq & [1, 10] & integer \\
LGBM & colsample\_bytree & [0.50, 1.00] & uniform \\
LGBM & reg\_alpha & [$10^{-8}$, 10] & log-uniform \\
LGBM & reg\_lambda & [$10^{-8}$, 10] & log-uniform \\
LGBM & min\_split\_gain & [$10^{-8}$, 1] & log-uniform \\
\midrule

XGB & n\_estimators & [300, 1500], step 100 & integer \\
XGB & learning\_rate & [$5 \times 10^{-3}$, 0.08 & log-uniform \\
XGB & max\_depth & [3, 12] & integer \\
XGB & min\_child\_weight & [1, 20] & log-uniform \\
XGB & gamma & [$10^{-8}$, 10] & log-uniform \\
XGB & subsample & [0.50, 1.00] & uniform \\
XGB & colsample\_bytree & [0.40, 1.00] & uniform \\
XGB & colsample\_bylevel & [0.40, 1.00] & uniform \\
XGB & reg\_alpha & [$10^{-8}$, 10] & log-uniform \\
XGB & reg\_lambda & [$10^{-8}$, 20] & log-uniform \\
XGB & grow\_policy & \{depthwise, lossguide\} & categorical \\
XGB & booster & gbtree & fixed \\
\bottomrule
\end{tabular}

\label{tab:hyperparameter_search_space}
\end{table}

\newpage

\begin{table}[!hbpt]
\centering
\caption{Selected variogram models and fitted parameters for Ordinary Kriging of $NO_2$in the Kriging interpolation baseline for each station.}
\label{tab:interpolation_nugget_sill_range}
    \begin{tabular}{lrrrrr}
\toprule
\textbf{Station} &
\textbf{Variogram} &
{\textbf{Partial sill}} &
{\textbf{Sill}} &
{\textbf{Range (km)}} &
{\textbf{Nugget}} \\
\midrule

Marylebone Road    & Spherical    & 31.3479    & 34.4045    & 1.8608    & 3.0566 \\
Camden Kerbside    & Spherical    & 10.2275    & 13.8756    & 2.2328    & 3.6481 \\
Putney High Street    & Exponential    & 0.0011    & 42.7250    & 0.0373    & 42.7239 \\
Oxford Street    & Spherical    & 22.0926    & 31.0504    & 5.0849    & 8.9578 \\
Euston Road    & Exponential    & 62.5974    & 66.6085    & 0.0355    & 4.0111 \\

\bottomrule
\end{tabular}
\end{table}

% =========================
% TABLE SM 4
% =========================
\begin{table}[!htbp]
\centering
\scriptsize
\setlength{\tabcolsep}{3pt}
\renewcommand{\arraystretch}{0.88}

\caption{Best-performing model and performance metrics by pollutant, station, and scenario.}
\resizebox{\textwidth}{!}{
\begin{tabular}{lll l rrrrrr}
\toprule
Pollutant & Station & Scenario & Best model & RMSE & MAE & $R^2$ \\
\midrule
    NO$_2$ & Marylebone Road & S1 & XGB & 10.22 $\pm$ 1.71 & 7.73 $\pm$ 1.23 & 0.33 $\pm$ 0.17 \\
    NO$_2$ & Marylebone Road & S1-noT & LGBM & 10.20 $\pm$ 1.93 & 7.73 $\pm$ 1.26 & 0.34 $\pm$ 0.14 \\
    NO$_2$ & Marylebone Road & S2 & XGB & 9.17 $\pm$ 1.30 & 6.85 $\pm$ 0.86 & 0.47 $\pm$ 0.07 \\
    NO$_2$ & Marylebone Road & S2-noT & ETR & 9.30 $\pm$ 1.57 & 7.02 $\pm$ 0.95 & 0.46 $\pm$ 0.09 \\
    NO$_2$ & Marylebone Road & S3 & LGBM & 8.01 $\pm$ 1.43 & 5.96 $\pm$ 1.06 & 0.59 $\pm$ 0.09 \\
    NO$_2$ & Marylebone Road & S3-noT & LGBM & 8.16 $\pm$ 1.78 & 5.92 $\pm$ 1.02 & 0.58 $\pm$ 0.10 \\
    NO$_2$ & Camden Kerbside & S1 & XGB & 9.27 $\pm$ 2.31 & 6.76 $\pm$ 1.43 & 0.39 $\pm$ 0.16 \\
    NO$_2$ & Camden Kerbside & S1-noT & XGB & 9.23 $\pm$ 2.06 & 6.68 $\pm$ 1.30 & 0.39 $\pm$ 0.16 \\
    NO$_2$ & Camden Kerbside & S2 & XGB & 7.37 $\pm$ 2.07 & 5.38 $\pm$ 1.28 & 0.62 $\pm$ 0.11 \\
    NO$_2$ & Camden Kerbside & S2-noT & LGBM & 7.48 $\pm$ 2.11 & 5.43 $\pm$ 1.41 & 0.60 $\pm$ 0.13 \\
    NO$_2$ & Camden Kerbside & S3 & LGBM & 6.82 $\pm$ 1.83 & 5.01 $\pm$ 1.03 & 0.68 $\pm$ 0.08 \\
    NO$_2$ & Camden Kerbside & S3-noT & LGBM & 6.80 $\pm$ 1.83 & 4.88 $\pm$ 1.01 & 0.68 $\pm$ 0.08 \\
    NO$_2$ & Putney High Street & S1 & XGB & 9.13 $\pm$ 1.01 & 6.99 $\pm$ 0.95 & 0.56 $\pm$ 0.05 \\
    NO$_2$ & Putney High Street & S1-noT & LGBM & 9.43 $\pm$ 1.14 & 7.22 $\pm$ 1.04 & 0.53 $\pm$ 0.05 \\
    NO$_2$ & Putney High Street & S2 & LGBM & 8.24 $\pm$ 1.57 & 6.20 $\pm$ 1.45 & 0.64 $\pm$ 0.11 \\
    NO$_2$ & Putney High Street & S2-noT & LGBM & 8.51 $\pm$ 1.85 & 6.33 $\pm$ 1.66 & 0.61 $\pm$ 0.13 \\
    NO$_2$ & Putney High Street & S3 & XGB & 7.65 $\pm$ 1.17 & 5.61 $\pm$ 1.02 & 0.69 $\pm$ 0.06 \\
    NO$_2$ & Putney High Street & S3-noT & LGBM & 7.81 $\pm$ 1.32 & 5.66 $\pm$ 1.13 & 0.68 $\pm$ 0.07 \\
    NO$_2$ & Oxford Street & S1 & XGB & 10.16 $\pm$ 2.35 & 7.83 $\pm$ 1.73 & 0.13 $\pm$ 0.15 \\
    NO$_2$ & Oxford Street & S1-noT & XGB & 10.44 $\pm$ 2.55 & 8.05 $\pm$ 1.79 & 0.08 $\pm$ 0.19 \\
    NO$_2$ & Oxford Street & S2 & LGBM & 7.90 $\pm$ 1.97 & 6.02 $\pm$ 1.58 & 0.47 $\pm$ 0.14 \\
    NO$_2$ & Oxford Street & S2-noT & LGBM & 8.09 $\pm$ 2.17 & 6.18 $\pm$ 1.73 & 0.44 $\pm$ 0.18 \\
    NO$_2$ & Oxford Street & S3 & XGB & 6.99 $\pm$ 1.58 & 5.33 $\pm$ 1.14 & 0.57 $\pm$ 0.16 \\
    NO$_2$ & Oxford Street & S3-noT & XGB & 7.40 $\pm$ 1.56 & 5.58 $\pm$ 1.12 & 0.50 $\pm$ 0.23 \\
    NO$_2$ & Euston Road & S1 & ETR & 18.53 $\pm$ 6.04 & 13.16 $\pm$ 2.78 & 0.09 $\pm$ 0.49 \\
    NO$_2$ & Euston Road & S1-noT & LGBM & 18.03 $\pm$ 6.92 & 12.32 $\pm$ 2.84 & 0.21 $\pm$ 0.24 \\
    NO$_2$ & Euston Road & S2 & XGB & 17.91 $\pm$ 5.91 & 12.33 $\pm$ 2.43 & 0.16 $\pm$ 0.43 \\
    NO$_2$ & Euston Road & S2-noT & RF & 17.30 $\pm$ 7.13 & 11.40 $\pm$ 2.91 & 0.28 $\pm$ 0.26 \\
    NO$_2$ & Euston Road & S3 & LGBM & 15.69 $\pm$ 6.84 & 9.96 $\pm$ 2.40 & 0.43 $\pm$ 0.17 \\
    NO$_2$ & Euston Road & S3-noT & RF & 15.75 $\pm$ 7.05 & 9.72 $\pm$ 2.57 & 0.43 $\pm$ 0.19 \\
    O$_3$ & Marylebone Road & S1 & LGBM & 10.65 $\pm$ 1.58 & 8.51 $\pm$ 1.17 & 0.25 $\pm$ 0.28 \\
    O$_3$ & Marylebone Road & S1-noT & LGBM & 11.09 $\pm$ 1.71 & 8.90 $\pm$ 1.34 & 0.15 $\pm$ 0.44 \\
    O$_3$ & Marylebone Road & S2 & LGBM & 7.50 $\pm$ 1.74 & 6.03 $\pm$ 1.57 & 0.64 $\pm$ 0.08 \\
    O$_3$ & Marylebone Road & S2-noT & LGBM & 7.85 $\pm$ 1.46 & 6.35 $\pm$ 1.34 & 0.59 $\pm$ 0.14 \\
    O$_3$ & Marylebone Road & S3 & XGB & 6.46 $\pm$ 1.73 & 5.23 $\pm$ 1.65 & 0.74 $\pm$ 0.06 \\
    O$_3$ & Marylebone Road & S3-noT & LGBM & 6.73 $\pm$ 1.59 & 5.38 $\pm$ 1.48 & 0.71 $\pm$ 0.06 \\
    PM$_{10}$ & Marylebone Road & S1 & RF & 7.21 $\pm$ 0.75 & 5.41 $\pm$ 0.52 & 0.13 $\pm$ 0.07 \\
    PM$_{10}$ & Marylebone Road & S1-noT & RF & 7.31 $\pm$ 0.78 & 5.46 $\pm$ 0.55 & 0.10 $\pm$ 0.06 \\
    PM$_{10}$ & Marylebone Road & S2 & LGBM & 5.55 $\pm$ 0.44 & 4.09 $\pm$ 0.41 & 0.48 $\pm$ 0.08 \\
    PM$_{10}$ & Marylebone Road & S2-noT & LGBM & 5.58 $\pm$ 0.37 & 4.11 $\pm$ 0.34 & 0.47 $\pm$ 0.08 \\
    PM$_{10}$ & Marylebone Road & S3 & XGB & 5.35 $\pm$ 0.25 & 4.03 $\pm$ 0.26 & 0.51 $\pm$ 0.10 \\
    PM$_{10}$ & Marylebone Road & S3-noT & LGBM & 5.40 $\pm$ 0.30 & 4.03 $\pm$ 0.29 & 0.50 $\pm$ 0.08 \\
    PM$_{10}$ & Camden Kerbside & S1 & XGB & 7.72 $\pm$ 1.33 & 5.52 $\pm$ 0.42 & 0.03 $\pm$ 0.07 \\
    PM$_{10}$ & Camden Kerbside & S1-noT & RF & 7.67 $\pm$ 1.41 & 5.42 $\pm$ 0.43 & 0.05 $\pm$ 0.06 \\
    PM$_{10}$ & Camden Kerbside & S2 & XGB & 5.24 $\pm$ 1.08 & 3.53 $\pm$ 0.35 & 0.56 $\pm$ 0.03 \\
    PM$_{10}$ & Camden Kerbside & S2-noT & LGBM & 5.21 $\pm$ 1.07 & 3.54 $\pm$ 0.34 & 0.56 $\pm$ 0.03 \\
    PM$_{10}$ & Camden Kerbside & S3 & XGB & 4.95 $\pm$ 0.92 & 3.40 $\pm$ 0.32 & 0.60 $\pm$ 0.02 \\
    PM$_{10}$ & Camden Kerbside & S3-noT & LGBM & 4.98 $\pm$ 0.93 & 3.42 $\pm$ 0.33 & 0.60 $\pm$ 0.02 \\
    PM$_{10}$ & Putney High Street & S1 & RF & 6.76 $\pm$ 1.01 & 5.13 $\pm$ 0.68 & 0.00 $\pm$ 0.14 \\
    PM$_{10}$ & Putney High Street & S1-noT & LGBM & 6.81 $\pm$ 0.99 & 5.16 $\pm$ 0.65 & -0.01 $\pm$ 0.15 \\
    PM$_{10}$ & Putney High Street & S2 & LGBM & 5.09 $\pm$ 1.55 & 3.74 $\pm$ 1.26 & 0.43 $\pm$ 0.25 \\
    PM$_{10}$ & Putney High Street & S2-noT & LGBM & 5.01 $\pm$ 1.54 & 3.64 $\pm$ 1.27 & 0.45 $\pm$ 0.21 \\
    PM$_{10}$ & Putney High Street & S3 & RF & 4.45 $\pm$ 1.73 & 3.27 $\pm$ 1.28 & 0.57 $\pm$ 0.21 \\
    PM$_{10}$ & Putney High Street & S3-noT & LGBM & 4.45 $\pm$ 1.68 & 3.25 $\pm$ 1.25 & 0.56 $\pm$ 0.22 \\
    PM$_{2.5}$ & Marylebone Road & S1 & XGB & 4.97 $\pm$ 1.42 & 3.48 $\pm$ 0.90 & -0.05 $\pm$ 0.11 \\
    PM$_{2.5}$ & Marylebone Road & S1-noT & LGBM & 5.02 $\pm$ 1.40 & 3.51 $\pm$ 0.91 & -0.08 $\pm$ 0.12 \\
    PM$_{2.5}$ & Marylebone Road & S2 & RF & 3.58 $\pm$ 0.81 & 2.54 $\pm$ 0.34 & 0.42 $\pm$ 0.18 \\
    PM$_{2.5}$ & Marylebone Road & S2-noT & ETR & 3.57 $\pm$ 0.87 & 2.51 $\pm$ 0.39 & 0.43 $\pm$ 0.16 \\
    PM$_{2.5}$ & Marylebone Road & S3 & RF & 3.46 $\pm$ 0.76 & 2.43 $\pm$ 0.24 & 0.45 $\pm$ 0.19 \\
    PM$_{2.5}$ & Marylebone Road & S3-noT & LGBM & 3.44 $\pm$ 0.81 & 2.42 $\pm$ 0.25 & 0.46 $\pm$ 0.19 \\
    PM$_{2.5}$ & Camden Kerbside & S1 & XGB & 5.18 $\pm$ 1.69 & 3.29 $\pm$ 0.39 & -0.11 $\pm$ 0.16 \\
    PM$_{2.5}$ & Camden Kerbside & S1-noT & XGB & 5.26 $\pm$ 1.72 & 3.36 $\pm$ 0.38 & -0.15 $\pm$ 0.19 \\
    PM$_{2.5}$ & Camden Kerbside & S2 & ETR & 3.48 $\pm$ 1.54 & 2.20 $\pm$ 0.25 & 0.50 $\pm$ 0.17 \\
    PM$_{2.5}$ & Camden Kerbside & S2-noT & ETR & 3.51 $\pm$ 1.56 & 2.21 $\pm$ 0.26 & 0.49 $\pm$ 0.17 \\
    PM$_{2.5}$ & Camden Kerbside & S3 & RF & 3.33 $\pm$ 1.53 & 2.11 $\pm$ 0.27 & 0.54 $\pm$ 0.17 \\
    PM$_{2.5}$ & Camden Kerbside & S3-noT & ETR & 3.38 $\pm$ 1.56 & 2.12 $\pm$ 0.26 & 0.53 $\pm$ 0.17 \\
\bottomrule
\end{tabular}%
}

\label{tab:best_model_metrics}
\end{table}

% =========================
% TABLE SM 5
% =========================
\clearpage

\begin{table}[htbp]
\caption{Percentage contribution (\%) of feature groups to the model predictions, grouped by pollutant, station, and scenario. N-S pollutants refer to the pollutants measured at the neighbouring stations from the corresponding scenario and the site to be estimated.}
\label{tab:Percentage_contribution}
\begin{tabular}{llllrrrr}
\toprule
Pollutant & Station & Scenario & Model & Temporal & Meteorological & Traffic & N-S pollutants \\
\midrule
NO$_2$ & Euston Road & Scen 1 & ETR & 36.58 & 37.14 & 26.28 & -- \\
NO$_2$ & Euston Road & Scen 2 & XGB & 18.62 & 44.95 & 25.57 & 10.85 \\
NO$_2$ & Euston Road & Scen 3 & LGBM & 9.08 & 22.54 & 13.97 & 54.42 \\
NO$_2$ & Camden Kerbside & Scen 1 & XGB & 19.64 & 55.78 & 24.58 & -- \\
NO$_2$ & Camden Kerbside & Scen 2 & XGB & 18.70 & 32.64 & 19.11 & 29.55 \\
NO$_2$ & Camden Kerbside & Scen 3 & LGBM & 11.02 & 17.16 & 13.05 & 58.77 \\
NO$_2$ & Marylebone Road & Scen 1 & XGB & 30.03 & 47.78 & 22.19 & -- \\
NO$_2$ & Marylebone Road & Scen 2 & XGB & 24.00 & 43.09 & 17.04 & 15.87 \\
NO$_2$ & Marylebone Road & Scen 3 & LGBM & 16.04 & 23.66 & 11.88 & 48.41 \\
NO$_2$ & Putney High Street & Scen 1 & XGB & 18.68 & 55.07 & 26.26 & -- \\
NO$_2$ & Putney High Street & Scen 2 & LGBM & 12.63 & 49.28 & 22.61 & 15.48 \\
NO$_2$ & Putney High Street & Scen 3 & XGB & 8.30 & 32.26 & 22.30 & 37.14 \\
NO$_2$ & Oxford Street & Scen 1 & XGB & 40.65 & 49.39 & 9.95 & -- \\
NO$_2$ & Oxford Street & Scen 2 & LGBM & 31.82 & 32.03 & 9.50 & 26.66 \\
NO$_2$ & Oxford Street & Scen 3 & XGB & 20.19 & 22.97 & 9.65 & 47.19 \\
O$_3$ & Marylebone Road & Scen 1 & LGBM & 23.14 & 66.34 & 10.51 & -- \\
O$_3$ & Marylebone Road & Scen 2 & LGBM & 14.02 & 41.44 & 9.04 & 35.50 \\
O$_3$ & Marylebone Road & Scen 3 & XGB & 12.33 & 24.00 & 5.82 & 57.85 \\
PM$_{10}$ & Camden Kerbside & Scen 1 & XGB & 22.85 & 59.72 & 17.42 & -- \\
PM$_{10}$ & Camden Kerbside & Scen 2 & XGB & 16.20 & 28.80 & 12.48 & 42.52 \\
PM$_{10}$ & Camden Kerbside & Scen 3 & XGB & 10.56 & 16.74 & 6.08 & 66.62 \\
PM$_{10}$ & Marylebone Road & Scen 1 & RF & 18.23 & 65.00 & 16.77 & -- \\
PM$_{10}$ & Marylebone Road & Scen 2 & LGBM & 11.72 & 36.28 & 7.42 & 44.57 \\
PM$_{10}$ & Marylebone Road & Scen 3 & XGB & 11.49 & 26.31 & 2.39 & 59.81 \\
PM$_{10}$ & Putney High Street & Scen 1 & RF & 30.96 & 57.62 & 11.42 & -- \\
PM$_{10}$ & Putney High Street & Scen 2 & LGBM & 15.99 & 39.74 & 4.38 & 39.90 \\
PM$_{10}$ & Putney High Street & Scen 3 & RF & 6.23 & 17.24 & 1.10 & 75.43 \\
PM$_{2.5}$ & Camden Kerbside & Scen 1 & XGB & 30.07 & 61.86 & 8.08 & -- \\
PM$_{2.5}$ & Camden Kerbside & Scen 2 & ETR & 18.58 & 11.22 & 6.87 & 63.33 \\
PM$_{2.5}$ & Camden Kerbside & Scen 3 & RF & 3.56 & 7.02 & 2.41 & 87.01 \\
PM$_{2.5}$ & Marylebone Road & Scen 1 & XGB & 34.50 & 58.18 & 7.31 & -- \\
PM$_{2.5}$ & Marylebone Road & Scen 2 & RF & 15.98 & 27.63 & 2.81 & 53.58 \\
PM$_{2.5}$ & Marylebone Road & Scen 3 & RF & 7.15 & 19.62 & 0.93 & 72.30 \\
\bottomrule
\end{tabular}
\end{table}

%\appendix

%\section{Supplementary Material}
%\input{supplementary_material.tex}

\end{document}